\documentclass[11pt]{article}

\usepackage[preprint]{acl}

\usepackage{times}
\usepackage{latexsym}

\usepackage[T1]{fontenc}

\usepackage[utf8]{inputenc}

\usepackage{microtype}

\usepackage{inconsolata}

\usepackage{graphicx}

\title{\textsc{OpenSeal}: Good, Fast, and Cheap Construction of \\ an Open-Source Southeast Asian LLM via Parallel Data}

\author{Tan Sang Nguyen,
    Muhammad Reza Qorib, 
    Hwee Tou Ng \\
    Department of Computer Science, National University of Singapore\\
  \texttt{e1583535@u.nus.edu, mrqorib@u.nus.edu, dcsnght@nus.edu.sg}}

\begin{document}
\maketitle
\begin{abstract}
Large language models (LLMs) have proven to be effective tools for a wide range of natural language processing (NLP) applications. Although many LLMs are multilingual, most remain English-centric and perform poorly on low-resource languages. Recently, several Southeast Asia–focused LLMs have been developed, but none are truly open source, as they do not publicly disclose their training data. Truly open-source models are important for transparency and for enabling a deeper and more precise understanding of LLM internals and development, including biases, generalization, and multilinguality. Motivated by recent advances demonstrating the effectiveness of parallel data in improving multilingual performance, we conduct controlled and comprehensive experiments to study the effectiveness of parallel data in continual pretraining of LLMs. Our findings show that using only parallel data is the most effective way to extend an LLM to new languages. Using just 34.7B tokens of parallel data and 180 hours on 8× NVIDIA H200 GPUs, we built \textsc{OpenSeal}, the first truly open Southeast Asian LLM that rivals the performance of existing models of similar size.
\end{abstract}

\section{Introduction}

With the democratization of large language model (LLM) development, enabled by open-source efforts from the research community, region-specific LLMs have been proposed to improve performance across particular language groups, especially low-resource languages. Consistent with the no free lunch theorem \cite{Wolpert1997}, numerous studies have shown that region-specific LLMs outperform one-size-fits-all models \cite{sengupta2023jaisjaischatarabiccentricfoundation, nguyen-etal-2024-seallms}.

The most efficient way to build a region-specific LLM is to continually train a model that has already been pretrained on trillions of tokens—standing on the shoulders of giants. Prior work has shown that this approach is not only more efficient but also more effective, achieving better performance than training from scratch \cite{zheng-etal-2024-breaking}. Most existing approaches adapt an LLM by further training it on a mixture of monolingual corpora corresponding to the target languages \cite{ng2025sealionsoutheastasianlanguages}.

Recently, increasing effort has been devoted to developing large language models for the Southeast Asian region. These efforts include compiling large-scale training corpora \cite{ng2025sealionsoutheastasianlanguages}, building evaluation benchmarks \cite{lovenia-etal-2024-seacrowd}, and training LLMs that support languages spoken in the region \cite{nguyen-etal-2024-seallms}. However, none of these models are fully open source. Existing Southeast Asian LLMs are open-weight models: while their model weights are publicly available, their training data (including the training data of their base LLMs) and training code are not fully disclosed.

Without complete transparency regarding their training data, such models risk being poisoned, either deliberately by the model provider or inadvertently through the inclusion of malicious content from the Internet. \citet{zhang2025persistent} show that poisoning as little as 0.1\% of the training data can induce harmful behavior in an LLM, and that this harm can persist even after additional preference tuning intended to improve safety and helpfulness. This makes non–fully open-source LLMs risky for security-sensitive applications, such as those in government institutions.

Recent work has demonstrated the effectiveness of parallel data in improving multilingual capabilities in LLMs \cite{qorib-etal-2025-just, fujii2024continual}. Building on this advancement, we conduct a controlled and comprehensive investigation into the effectiveness of parallel data for extending LLMs to new languages. Based on this investigation, we develop \textsc{OpenSeal} (\underline{Open}-source \underline{S}outh\underline{E}ast \underline{A}sian \underline{L}LM), a fully open-source Southeast Asian LLM. \textsc{OpenSeal} is trained using only 34.7B tokens of parallel data and 180 hours on 8× NVIDIA H200 GPUs. The training cost based on commercial rates charged by cloud providers\footnote{For example, \href{https://aws.amazon.com/ec2/pricing/on-demand/}{AWS \texttt{p5en.48xlarge}}} is under US\$12{,}000.

To summarize, our contributions are as follows:
\begin{itemize}
    \item We propose a good, fast, and cheap way of adapting an LLM to new languages by utilizing parallel data. We show that under the same token budget, utilizing only parallel data is the most efficient. To the best of our knowledge, we are the first to present a comprehensive controlled investigation of the use of parallel data for continual pretraining.
    \item We built a new Southeast Asian LLM, \textsc{OpenSeal}, by continually training an open-source English-only LLM (OLMo 2; \citealt{walsh2025}) on a relatively small amount of parallel data. On translation and commonsense reasoning benchmarks, \textsc{OpenSeal} performs at least as well as --- and in some cases outperforms --- existing Southeast Asian LLMs that were based on high-performing multilingual LLMs and trained on a much larger amount of multilingual data.
    \item Our model is fully open (open source, open data, and open weight\footnote{Source code, training data, and model weights will be released upon publication.}), which allows future research to investigate the multilinguality of LLMs in a more precise and detailed manner. To the best of our knowledge, our model is the first fully open Southeast Asian LLM.
\end{itemize}

\section{Related Work}
In this section, we will discuss recent work that investigates the effectiveness of parallel data for LLM training---pretraining, continual pretraining, and instruction tuning---and recent efforts in building LLMs for the Southeast Asian region.

\subsection{Parallel Data in LLM training}
Parallel data have been commonly used in training encoder-only \cite{ConneauL19, ouyang-etal-2021-ernie} and encoder–decoder \cite{liu-etal-2020-multilingual-denoising, chi-etal-2021-mt6} LLMs, but early decoder-only LLMs \cite{bloom} did not utilize parallel data and were trained solely on a mixture of monolingual data.

\citet{qorib-etal-2025-just} conduct a comprehensive investigation into the effects of parallel data in LLM pretraining. They compare pretraining with and without parallel data in a controlled setting by fixing the choice and order of the training data. Their experiments on a 1B-parameter model trained on 167B tokens show that incorporating parallel data improves multilingual capabilities more than monolingual data, especially when parallel data are placed at the end of training. While these findings highlight the value of parallel data in LLM training, the experiments are limited to 1B models trained on a relatively small number of tokens compared to production-scale LLMs.

A recent multilingual LLM, Apertus \cite{apertus2025apertusdemocratizingopencompliant}, includes parallel data in one of its pretraining stages. Similarly, the model is trained with parallel data in the final (fifth) pretraining stage. Although the models have up to 70B parameters, the authors do not investigate the specific effect of parallel data during pretraining.

\citet{fujii2024continual} utilize parallel data to build a Japanese LLM by performing continual pretraining (CPT) of Llama 2 \cite{llama2}. They first train Llama 2 on 22 million parallel sentence pairs before further training it on monolingual data. The motivation for this training order is the assumption that parallel sentences can facilitate the transition from English-centric to Japanese-centric training.

In contrast, ALMA \cite{xu2024a} adapts Llama 2 for machine translation by training the model on parallel data only after it has been trained on monolingual data. The authors argue that monolingual training is necessary to improve the model’s proficiency in non-English languages before introducing parallel data.

These two approaches adopt the exact opposite order for utilizing parallel data, and this difference in training order is not inconsequential. We find that such differences can result in models with differing capabilities. Due to the lack of comprehensive empirical studies on the effects of parallel data in CPT, it remains unclear how parallel data influence continual pretraining and how they can be maximally leveraged to extend LLMs to new languages.

\citet{ranaldi-etal-2024-empowering} utilize parallel data as an additional signal during instruction tuning. Specifically, they employ a machine translation model to translate instruction-following datasets into multiple languages and incorporate the resulting translation pairs during instruction tuning. They find that this approach promotes better cross-lingual transfer. While instruction tuning is outside the scope of our current work, this method is orthogonal to ours and can be applied sequentially.

Our work extends the investigation of parallel data by \citet{qorib-etal-2025-just} to continual pretraining of a base LLM (OLMo 2) that is pretrained on more than four trillion tokens. In addition to studying CPT on a much larger scale, we scale the model size to 7B parameters. Although parallel data have been used previously for CPT, to the best of our knowledge, no comprehensive empirical study has systematically evaluated the role of parallel data in continual pretraining. Our work aims to fill this gap and demonstrates that parallel data are highly effective for extending LLMs to new languages.

\subsection{Southeast Asian LLMs}
Notable LLMs developed for the Southeast Asian region include SEA-LION v3.5~\cite{ng2025sealionsoutheastasianlanguages}, Sailor 2~\cite{dou-etal-2024-sailor}, and SeaLLM 3~\cite{zhang-etal-2025-seallms}. All of them were built on multilingual open-weight LLMs by performing continual pre-training (CPT) on Southeast Asian data, making the models more attuned to native texts from the region. Adapting high-performing multilingual models allows them to retain strong capabilities in English and Chinese, while also providing a significant head start on high-resource Southeast Asian languages such as Indonesian and Thai. In this paper, we will refer to languages using their ISO 639 codes, as listed in Table~\ref{tab:sta-sea_en-parallel_corpus}.

SEA-LION v3.5 has two versions: one built by performing CPT on Llama 2 8B and the other on Gemma 2 9B\footnote{At the time of writing, only the Llama-based version is publicly available.}. CPT was applied to the instruction-tuned versions of the models using 200B tokens of data, comprising 50B tokens of English, 110B tokens from 10 Southeast Asian languages (\texttt{id}, \texttt{km}, \texttt{lo}, \texttt{ms}, \texttt{my}, \texttt{ta}, \texttt{th}, \texttt{tl}, \texttt{vi}, \texttt{zh}), and 40B tokens of code. The models then underwent instruction tuning and preference tuning on both English and Southeast Asian data. Between training stages, a series of model-merging steps was employed to mitigate catastrophic forgetting. On eight nodes equipped with 8× NVIDIA H200 GPUs per node, the total training time was six days for the Llama-based version and ten days for the Gemma-based version.

A newer version of SEA-LION (v4) has been released on HuggingFace\footnote{\url{https://huggingface.co/collections/aisingapore/sea-lion-v4}}, built by performing CPT on Gemma 3 and Qwen 3. The Gemma version has 27B parameters, while the Qwen version has 32B parameters. The Gemma model was continually pretrained on approximately 500B tokens from 11 Southeast Asian languages (\texttt{en}, \texttt{id}, \texttt{km}, \texttt{lo}, \texttt{ms}, \texttt{my}, \texttt{ta}, \texttt{th}, \texttt{tl}, \texttt{vi}, \texttt{zh}), whereas the Qwen model was continually pretrained on approximately 100B tokens from 7 Southeast Asian languages (\texttt{id}, \texttt{ms}, \texttt{my}, \texttt{ta}, \texttt{th}, \texttt{tl}, \texttt{vi}). SEA-LION v4 also includes vision–language model variants adapted from Qwen 4B, Qwen 8B, and Gemma 27B. At the time of writing, no technical report or academic paper on SEA-LION v4 has been released.

Sailor 2 was built by expanding Qwen 2.5 using an approach similar to LlamaPro~\cite{wu-etal-2024-llama}, increasing the parameter counts from 0.5B to 1B, from 7B to 8B, and from 14B to 20B. These expansions were designed to mitigate potential catastrophic forgetting on English and Chinese tasks. The models were trained on 500B tokens of data, comprising 400B tokens from 13 Southeast Asian languages (\texttt{ceb}, \texttt{id}, \texttt{ilo}, \texttt{jv}, \texttt{km}, \texttt{lo}, \texttt{ms}, \texttt{my}, \texttt{su}, \texttt{th}, \texttt{tl}, \texttt{vi}, \texttt{war}) and 100B tokens of English text. CPT was performed in two stages. In the first stage, the model was trained on 450B tokens from a comprehensive dataset covering 8 Southeast Asian languages and English. In the second stage, it was trained on high-quality data covering 13 Southeast Asian languages along with English instruction-tuning data. The model then underwent two stages of instruction tuning and two stages of preference tuning.

SeaLLM 3 was built by performing CPT on Qwen 2 using data from 12 Southeast Asian languages (\texttt{en}, \texttt{id}, \texttt{jv}, \texttt{km}, \texttt{lo}, \texttt{ms}, \texttt{my}, \texttt{ta}, \texttt{th}, \texttt{tl}, \texttt{vi}, \texttt{zh}). Unlike the previously discussed Southeast Asian LLMs, SeaLLM updates only a small subset of model parameters identified as language-specific neurons. In addition, CPT was performed for one language at a time, resulting in multiple monolingual model variants, each specialized in a single Southeast Asian language. Subsequently, the monolingual models were merged into a single multilingual model. This training approach requires less data\footnote{The amount of training data used for SeaLLM is not disclosed.}.

\begin{figure*}[t]
\centering
\includegraphics[width=0.9\textwidth]{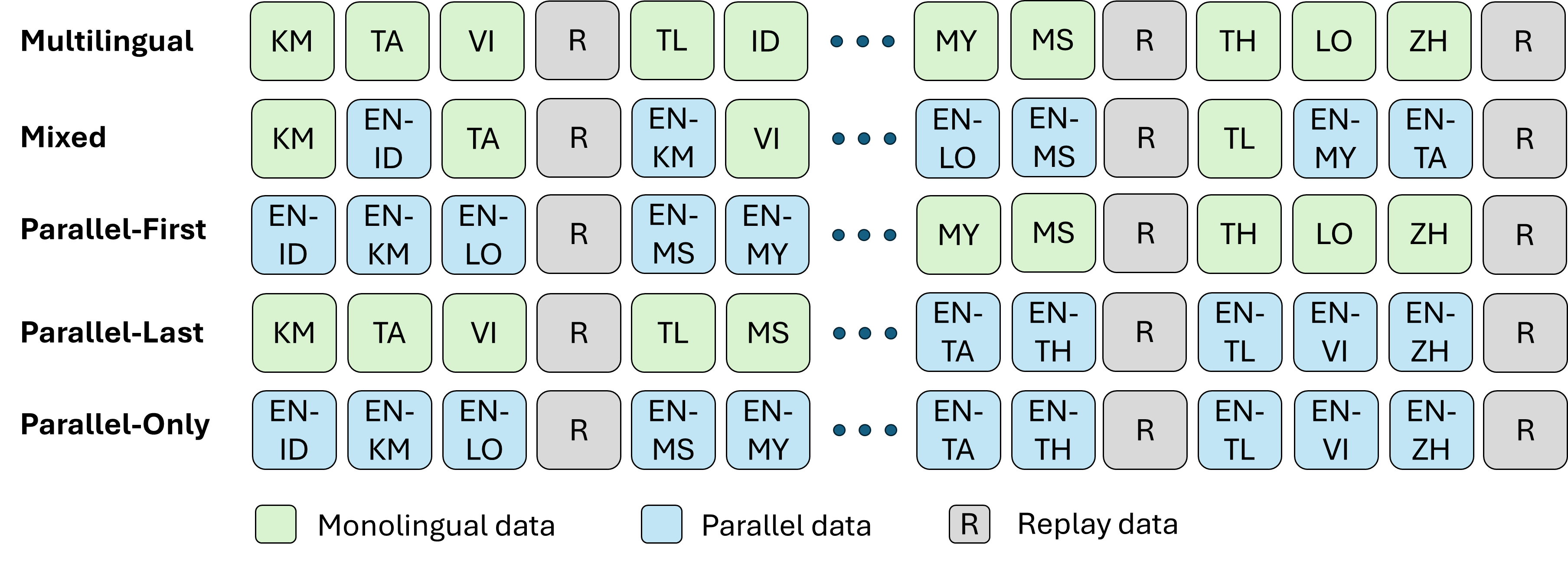}
\caption{Illustration of the training data sequences of our parallel data investigation.}
\label{fig:method}
\end{figure*}

\section{Methods}
\label{sec:methods}
To evaluate the effectiveness of parallel data, we vary the inclusion and placement of parallel data and try to maintain the choice and order of the other training data, while ensuring each data type is still uniformly distributed. To this end, we define each data source as small blocks of 262,144 tokens\footnote{This corresponds to 64 sequences of 4,096 tokens (the maximum context window size).} (Figure~\ref{fig:method}). Note that the order of blocks within a batch (eight blocks for the 1B model or sixteen blocks for the 7B model) is randomized.

Monolingual data blocks are constructed from a collection of texts in a single language, whereas parallel data blocks are formed by concatenating an English sentence and its Southeast Asian translation using the format ``\texttt{{source language}: {source sentence}\textbackslash n{target language}: {target sentence}}<|endoftext|>''. This format mimics incidental parallel data found in PaLM’s~\cite{10.5555/3648699.3648939} training set, as reported by~\citet{briakou-etal-2023-searching}. To ensure exposure to both translation directions, we randomize the order of the source and target languages within each block (e.g., for \texttt{en-id} blocks, some segments begin with English, while others begin with Indonesian).

Replay is crucial for mitigating catastrophic forgetting: even a small replay ratio (e.g., 5\%) yields significant advantages over using no replay. The replay proportion represents a trade-off between preserving proficiency in previously acquired domains or languages and adapting to new ones. \citet{ibrahim2024simplescalablestrategiescontinually} empirically analyze replay ratios and recommend a 25\% replay rate. Following this recommendation, we ensure that one out of every four blocks in our training data is replay data across all experimental settings.

We define five experimental settings: \textsc{Multilingual}, \textsc{Mixed}, \textsc{Parallel First}, \textsc{Parallel Last}, and \textsc{Parallel Only}. In the \textsc{Mixed}, \textsc{Parallel First}, and \textsc{Parallel Last} settings, we maintain an equal ratio of monolingual and parallel data blocks.

\subsection {Multilingual}
The \textsc{Multilingual} setting reflects the standard approach to extending an LLM to new languages, in which the model is continually pretrained solely on a mixture of monolingual corpora. We sample data blocks uniformly across all languages to prevent dominance by any single language. Continual pretraining on a mixture of monolingual corpora is similar to the approach used by SEA-LION and other Southeast Asian LLMs.

\subsection {Mixed}
In the \textsc{Mixed} setting, monolingual and parallel data blocks are uniformly interleaved. To ensure that each batch contains both data types, we enforce the presence of at least one monolingual block and one parallel block between two replay blocks. Mixing monolingual data and parallel data is also employed by Apertus in its fifth pretraining stage.

\subsection {Parallel First}
The \textsc{Parallel First} setting prioritizes parallel data in the early stages of continual pretraining. Once the parallel blocks are exhausted, training proceeds with multilingual data. This setting is motivated by the hypothesis that early exposure to parallel data facilitates the learning of cross-lingual alignments, which can subsequently be reinforced and generalized through multilingual training. Placing parallel data before monolingual data is also the CPT strategy adopted by \cite{fujii2024continual}.

\subsection {Parallel Last}
This setting begins with monolingual blocks and switches to parallel data after the monolingual blocks are exhausted. It is motivated by the hypothesis that extensive monolingual training improves general fluency in new languages before introducing aligned data for enhanced cross-lingual transfer. Placing monolingual data before parallel data is the CPT strategy used by ALMA.

\subsection {Parallel Only}
In the \textsc{Parallel Only} setting, continual pretraining relies exclusively on parallel data and replay data, entirely omitting monolingual data. This setting tests whether parallel data alone are sufficient—and potentially superior—to monolingual data for improving multilingual capabilities and cross-lingual transfer, and whether it should be exclusively used for CPT when available.

\section{Experiments}

\subsection{Model}
We build our models via continual pretraining (CPT) of OLMo 2, a fully open, decoder-only transformer language model. We experiment with the 1B and 7B OLMo 2 variants. The 1B model has 16 transformer layers, while the 7B model has 32 layers. OLMo 2 employs an autoregressive architecture designed for stable large-scale training, incorporating bias-free linear layers, rotary positional embeddings, RMS normalization, and SwiGLU activation. We first train the 1B model on 10B tokens to identify the optimal CPT strategy. We then scale the best configuration to the 7B model by training on 34.7B tokens and compare it against the \textsc{Multilingual} setting, which is the most commonly used CPT setting in prior work.

We keep the OLMo 2 architecture fixed and focus exclusively on the effects of training data composition and ordering during CPT. Starting from an OLMo 2 checkpoint pretrained on four trillion tokens of general-domain data, we further adapt the model using different mixtures of parallel, multilingual, and replay data discussed in Section \ref{sec:methods}. Aside from the data mixture, all training hyperparameters and model components are identical across the different settings, enabling a controlled comparison of how parallel and multilingual data interact under CPT. Following prior work, we use the WSD scheduler \cite{hu2024minicpm}. Details of the training hyperparameters are provided in Appendix \ref{sec:training_details}.

\subsection{Data}

\begin{table}
  \centering
  \begin{tabular}{llrrr}
    \hline
    \textbf{ISO} & \textbf{Language} & \textbf{\# sent.} & \multicolumn{2}{c}{\textbf{\# tokens}} \\
    \textbf{Code} & & & \textbf{SEA} & \textbf{EN} \\
    \hline
    \texttt{id} & Indonesian & 70.5M & 2.2B & 1.9B \\
    \texttt{km} & Khmer      & 5.8M  & 0.6B & 0.1B \\
    \texttt{lo} & Lao        & 4.2M  & 0.4B & 0.1B \\
    \texttt{ms} & Malay      & 56.8M & 1.2B & 1.0B \\
    \texttt{my} & Burmese    & 10.0M & 1.2B & 0.2B \\
    \texttt{ta} & Tamil      & 42.5M & 3.7B & 0.7B \\
    \texttt{th} & Thai       & 29.0M & 1.6B & 0.6B \\
    \texttt{tl} & Tagalog    & 63.6M & 1.4B & 1.2B \\
    \texttt{vi} & Vietnamese & 50.1M & 2.4B & 1.3B \\
    \texttt{zh} & Chinese    & 71.3M & 2.5B & 1.7B \\
    \hline
    Total &         & 403.8M & 17.2B & 8.8B \\
    \hline
  \end{tabular}
  \caption{Number of sentence pairs (\# sent.) and tokens (measured using the OLMo 2 tokenizer) of our parallel data.}
  \label{tab:sta-sea_en-parallel_corpus}
\end{table}

For the parallel data, we construct a large-scale SEA–English corpus covering multiple Southeast Asian languages, comprising approximately 403.8 million parallel sentence pairs in total (Table~\ref{tab:sta-sea_en-parallel_corpus}). The corpus contains 17.2 billion tokens in ten Southeast Asian languages and 8.8 billion tokens in English, as measured using the OLMo 2 tokenizer. Wherever possible, we rely on NLLB \cite{nllb2022} as the primary source of parallel data to ensure consistency across languages. For Thai, however, NLLB data are not available; therefore, we aggregate parallel data from multiple publicly available sources\footnote{bible-uedin, CCAligned, ELRC\_2922, GNOME, HPLT, KDE4, OpenSubtitles, QED, Tanzil, TED2020.}.

\begin{table*}
\centering
\begin{tabular}{ll|ccc|c}
\hline
\textbf{Model} & \textbf{Param} & \textbf{EN $\rightarrow$ XX} & \textbf{XX $\rightarrow$ EN} & \textbf{XNLI} & \textbf{Avg} \\
\hline
Multilingual & 1B & 12.51 & 14.50 & 43.28 & 23.43 \\
Mixed & 1B & \underline{23.69} & 25.63 & 43.28 & \underline{30.87} \\
Parallel First & 1B & 19.00 & 20.85 & \textbf{44.81} & 28.22 \\
Parallel Last & 1B & 22.69 & \textbf{26.55} & 42.84 & 30.69 \\
Parallel Only & 1B & \textbf{24.12} & \underline{26.24} & \underline{44.09} & \textbf{31.48} \\
\hline
\end{tabular}
\caption{\label{tab:1B_models_10B-AvgBLEU-AvgXNLI} Comparison of translation (BLEU scores) and commonsense reasoning performance under different 1B training settings. The highest score is bolded, while the second-highest score is underlined.}
\end{table*}

For the Southeast Asian multilingual data, we use SEA-PILE-v2~\cite{ng2025sealionsoutheastasianlanguages}, a large-scale multilingual corpus covering Southeast Asian languages. Since Chinese is not included in SEA-PILE-v2, we source Chinese monolingual data from the MAP-CC corpus~\cite{du2024chinesetinyllmpretraining}, which consists of high-quality text from diverse sources, including books, encyclopedias, academic papers, and web articles.

For the replay data, we sample from OLMo 2’s second-stage pretraining data, which consist of a diverse mixture of high-quality English corpora. They include filtered web data, instruction-tuned data, technical question–answer data, academic papers, encyclopedic text, and mathematics-focused data.

When sampling the data, we aim to sample each data source uniformly by equalizing the number of tokens drawn from each source. The total number of available tokens for each data source is provided in Appendix~\ref{sec:resources}.

\subsection{Evaluation}

We evaluate the models on translation and multilingual commonsense reasoning tasks. Translation quality is assessed using the BLEU metric \cite{papineni-etal-2002-bleu}, computed with SacreBLEU \cite{post-2018-call}. Translation performance is evaluated in both directions—from Southeast Asian languages to English and from English to Southeast Asian languages—using test sets from FLORES-200 \cite{nllb2022}. For all translation evaluations, we employ a fixed 5-shot prompting setup across all models reported in this study.

In addition to translation, we evaluate multilingual commonsense reasoning using XNLI \cite{conneau-etal-2018-xnli}, XCOPA \cite{ponti-etal-2020-xcopa}, and PAWS-X \cite{yang-etal-2019-paws}. All experiments are conducted using the lm-evaluation-harness\footnote{https://github.com/EleutherAI/lm-evaluation-harness}. For XNLI, we report results for English, Thai, Vietnamese, and Chinese. XCOPA is evaluated on English, Indonesian, Tamil, Thai, Vietnamese, and Chinese, covering a diverse set of typologically distinct languages. For PAWS-X, we evaluate performance on English and Chinese. This multilingual evaluation setup enables a consistent assessment of commonsense reasoning across languages and tasks. In all experiments, we assess statistical significance using paired bootstrap resampling \cite{koehn-2004-statistical} with 1,000 samples.

\section{Results}

\begin{table*}
\centering
\begin{tabular}{ll|cc|ccc}
\hline
\textbf{Model} & \textbf{Param} & 
\multicolumn{2}{c|}{\textbf{Translation}} & \multicolumn{3}{c}{\textbf{Reasoning}} 
 \\
 &  & \textbf{EN $\rightarrow$ XX} & \textbf{XX $\rightarrow$ EN} & \textbf{XNLI} & \textbf{XCOPA} & \textbf{PAWS-X} \\
\hline
Multilingual & 7B & 24.17 & 30.15 & \underline{45.73} & 70.30 & 60.08 \\
Parallel Only (\textsc{OpenSeal}) & 7B & \textbf{31.12} & \textbf{38.04} & 45.40 & 70.20 & \textbf{64.70} \\
\hline
SeaLLM v3 & 7B & 17.85 & 25.71 & 43.28 & \underline{70.53} & 60.18 \\
Sailor2 & 8B & \underline{31.09} & \underline{36.37} & 44.59 & \textbf{74.50} & \underline{64.53} \\
SEA-LION v3.5 & 8B & 25.37 & 32.03 & \textbf{46.11} & 70.37 & 63.58 \\
\hline
\end{tabular}
\caption{\label{tab:7B_models-translation-reasoning} Comparison of translation and commonsense reasoning performance across 7B–8B multilingual models. The highest score is bolded, while the second-highest score is underlined.}
\end{table*}

Our controlled experiments on the effect of parallel data show that the \textsc{Parallel Only} strategy is the most effective for extending LLMs to new languages. It significantly outperforms ($p < 0.01$) other settings in both translation directions, except when compared with \textsc{Parallel Last} in the XX $\rightarrow$ EN direction, where no statistically significant difference is observed (Table~\ref{tab:1B_models_10B-AvgBLEU-AvgXNLI}). In addition, \textsc{Parallel Only} significantly outperforms \textsc{Parallel Last} on XNLI. As no other configuration significantly outperforms \textsc{Parallel Only}, this strategy is superior overall.

Given the strong performance of \textsc{Parallel Only} at the 1B scale, we next examine whether this advantage persists as we scale the model size and training data. Table~\ref{tab:7B_models-translation-reasoning} reports results for the larger 7B model, trained on 34.7B tokens. Consistent with the 1B-scale findings, \textsc{Parallel Only} (\textsc{OpenSeal}) achieves substantial gains in translation quality, attaining the highest BLEU scores in both EN$\rightarrow$XX and XX$\rightarrow$EN directions. Improvements of 
\textsc{Parallel Only} (\textsc{OpenSeal}) over other Southeast Asian LLMs are statistically significant ($p < 0.01$), except when compared with Sailor 2 in the EN $\rightarrow$ XX direction, where no significant difference is observed.

Beyond translation, \textsc{Parallel Only} (\textsc{OpenSeal}) remains competitive on XNLI, XCOPA, and PAWS-X. It achieves the best score on PAWS-X among all compared models, significantly outperforming SeaLLM v3 and \textsc{Multilingual} and showing no significant differences on XNLI and XCOPA, except when compared with Sailor 2 on XCOPA.

Overall, across all evaluated tasks and models, \textsc{Parallel Only} (\textsc{OpenSeal}) is significantly outperformed only by Sailor 2 on XCOPA, while it significantly outperforms all models—including Sailor 2—on the translation task. This result is notable because other Southeast Asian LLMs are built from strong multilingual LLMs and continually trained on substantially larger corpora (400B tokens for Sailor 2 and 200B tokens for SEA-LION v3.5), whereas \textsc{OpenSeal} is built by continual pretraining of an English-only LLM on just 34.7B tokens. These findings highlight the effectiveness and efficiency of parallel data for CPT.

\section{Analysis}

\subsection{Impact of Parallel Data}
To isolate the contribution of parallel SEA sentences, we introduce an ablated variant of \textsc{Parallel Only}, denoted as \textsc{Multilingual Replacement}, in which the SEA sentences in the parallel data blocks are replaced with multilingual SEA text from SEA-PILE-v2. All other components are kept identical: the replay data, the English-side data, and the adjacency-based training format remain unchanged. This design allows us to assess whether the advantages of \textsc{Parallel Only} stem from bilingual alignment enabled by genuine parallel sentences.

Table~\ref{tab:7B_parallel-multi_non_u-translation-reasoning} shows that this substitution results in significant ($p < 0.05$) degradation in translation performance and PAWS-X. Specifically, the EN~$\rightarrow$~XX translation score drops from 31.12 to 22.03, the XX~$\rightarrow$~EN score from 38.04 to 23.19, and the PAWS-X score from 64.70 to 62.45. In contrast, no significant differences are observed on XNLI or XCOPA.

Overall, these findings directly address our primary research question, demonstrating that the benefits of \textsc{Parallel Only} arise from deliberate bilingual alignment provided by authentic parallel sentences. The results further indicate that performing CPT using only parallel data does not cause the model to overfit to translation, as it also yields improved performance on commonsense reasoning tasks.

\begin{table*}
\centering
\begin{tabular}{ll|cc|ccc}
\hline
\textbf{Model} & \textbf{Param} & 
\multicolumn{2}{c|}{\textbf{Translation}} & \multicolumn{3}{c}{\textbf{Reasoning}} 
 \\
 &  & \textbf{EN $\rightarrow$ XX} & \textbf{XX $\rightarrow$ EN} & \textbf{XNLI} & \textbf{XCOPA} & \textbf{PAWS-X} \\
\hline
Parallel Only & 7B & \textbf{31.12} & \textbf{38.04} & 45.40 & \textbf{70.20} & \textbf{64.70} \\
Multilingual Replacement & 7B & 22.03& 23.19 & \textbf{46.13} & 69.83 & 62.45 \\
\hline
\end{tabular}
\caption{\label{tab:7B_parallel-multi_non_u-translation-reasoning} Comparison of translation and commonsense reasoning performance between \textsc{Parallel Only} and \textsc{Multilingual Replacement}. The highest score is bolded.}
\end{table*}

\begin{figure*}[h!]
  \centering
  \begin{minipage}{0.2\textwidth}
    \centering
    \includegraphics[width=\linewidth]{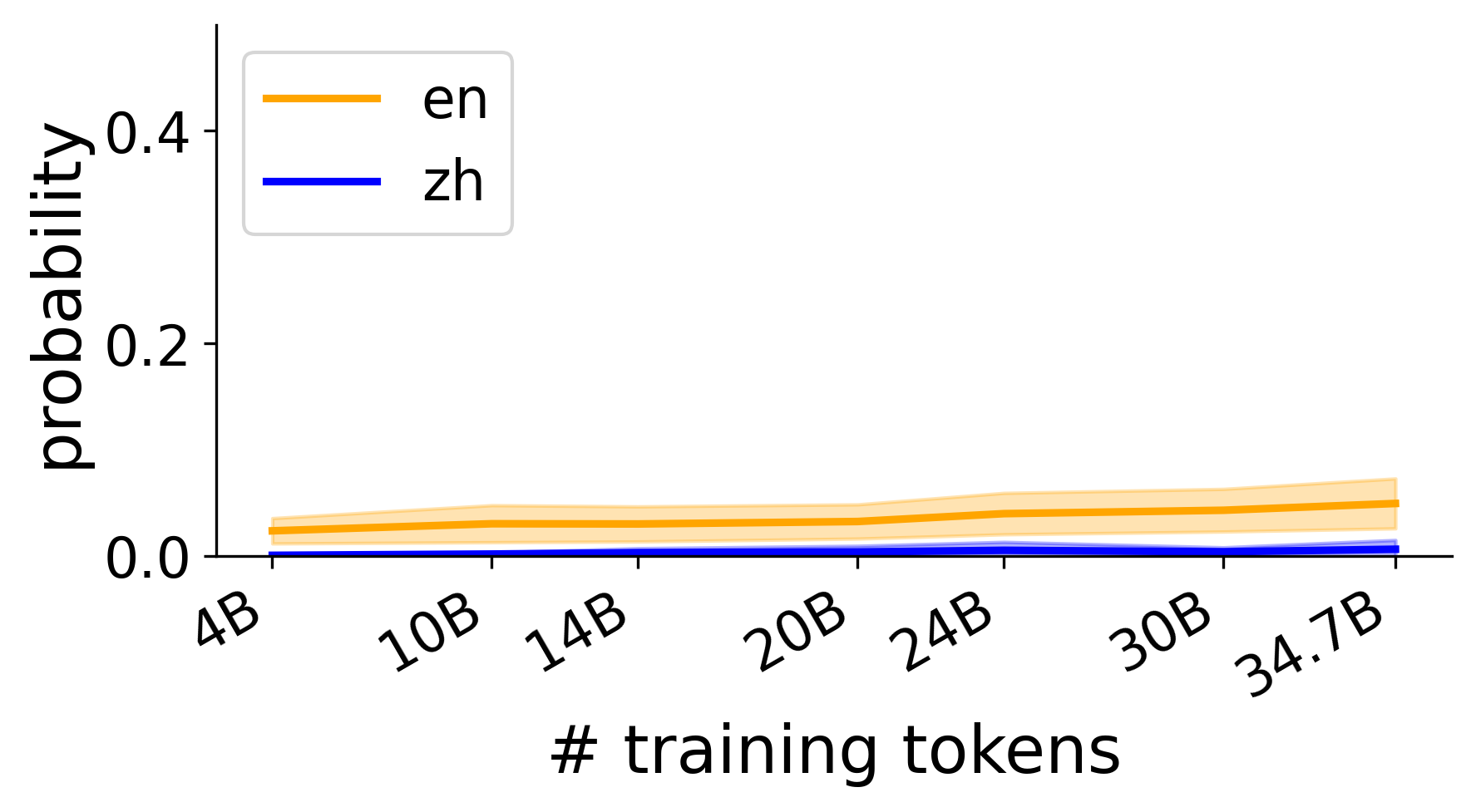}\\
    \small (a) Layer 20
  \end{minipage}\hfill
  \begin{minipage}{0.2\textwidth}
    \centering
    \includegraphics[width=\linewidth]{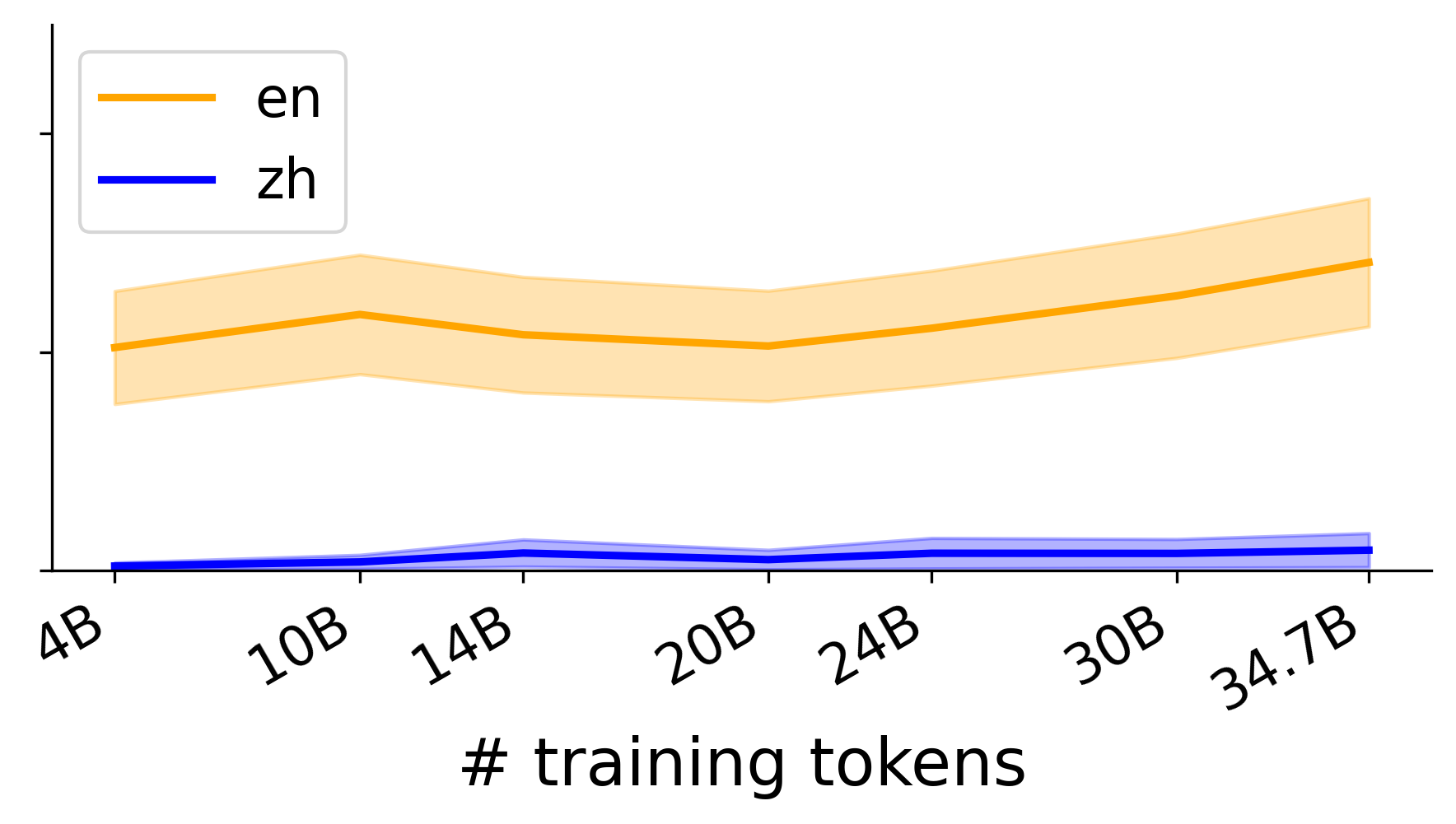}\\
    \small (b) Layer 24
  \end{minipage}\hfill
  \begin{minipage}{0.2\textwidth}
    \centering
    \includegraphics[width=\linewidth]{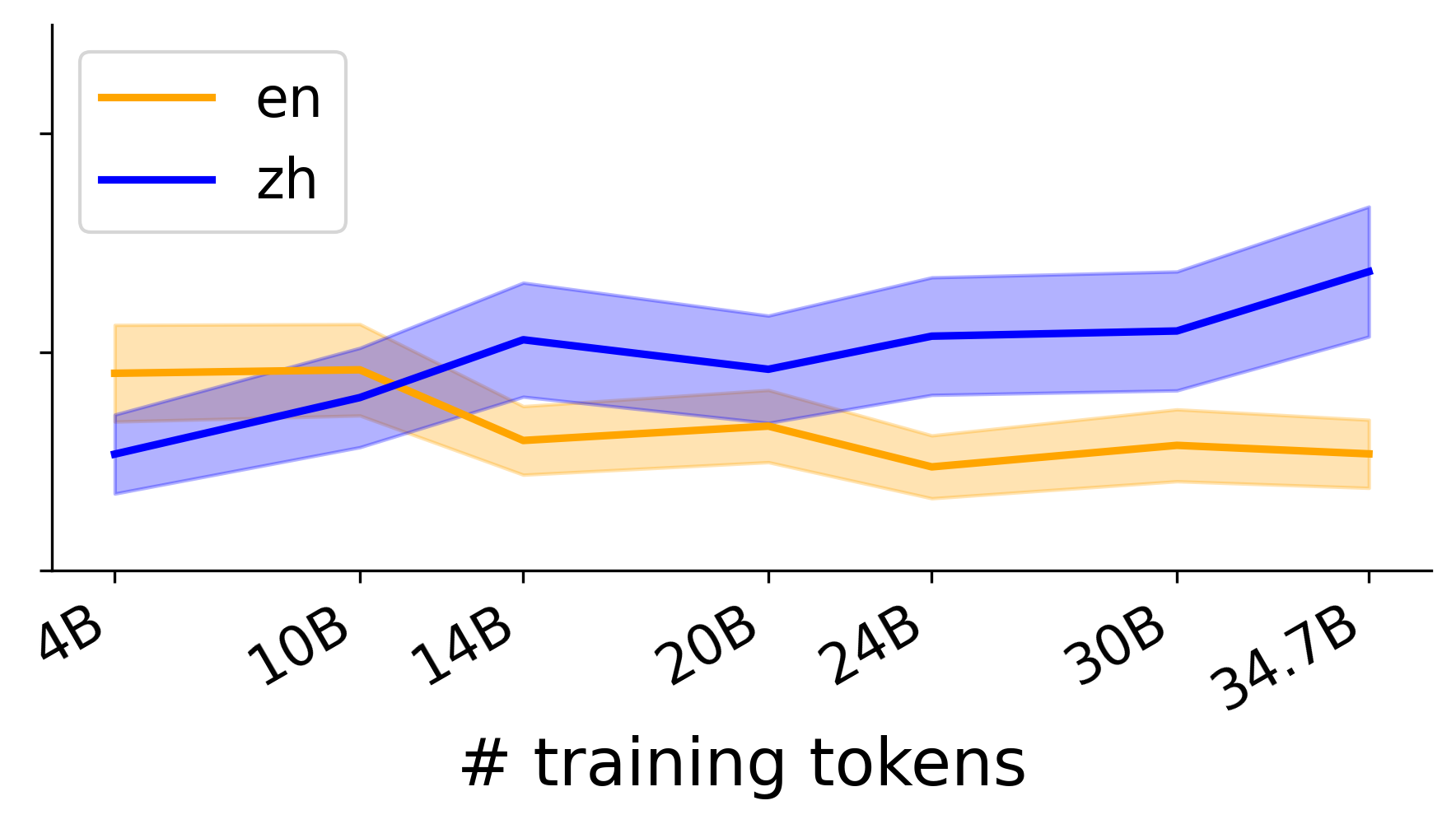}\\
    \small (c) Layer 30
  \end{minipage}\hfill
  \begin{minipage}{0.2\textwidth}
    \centering
    \includegraphics[width=\linewidth]{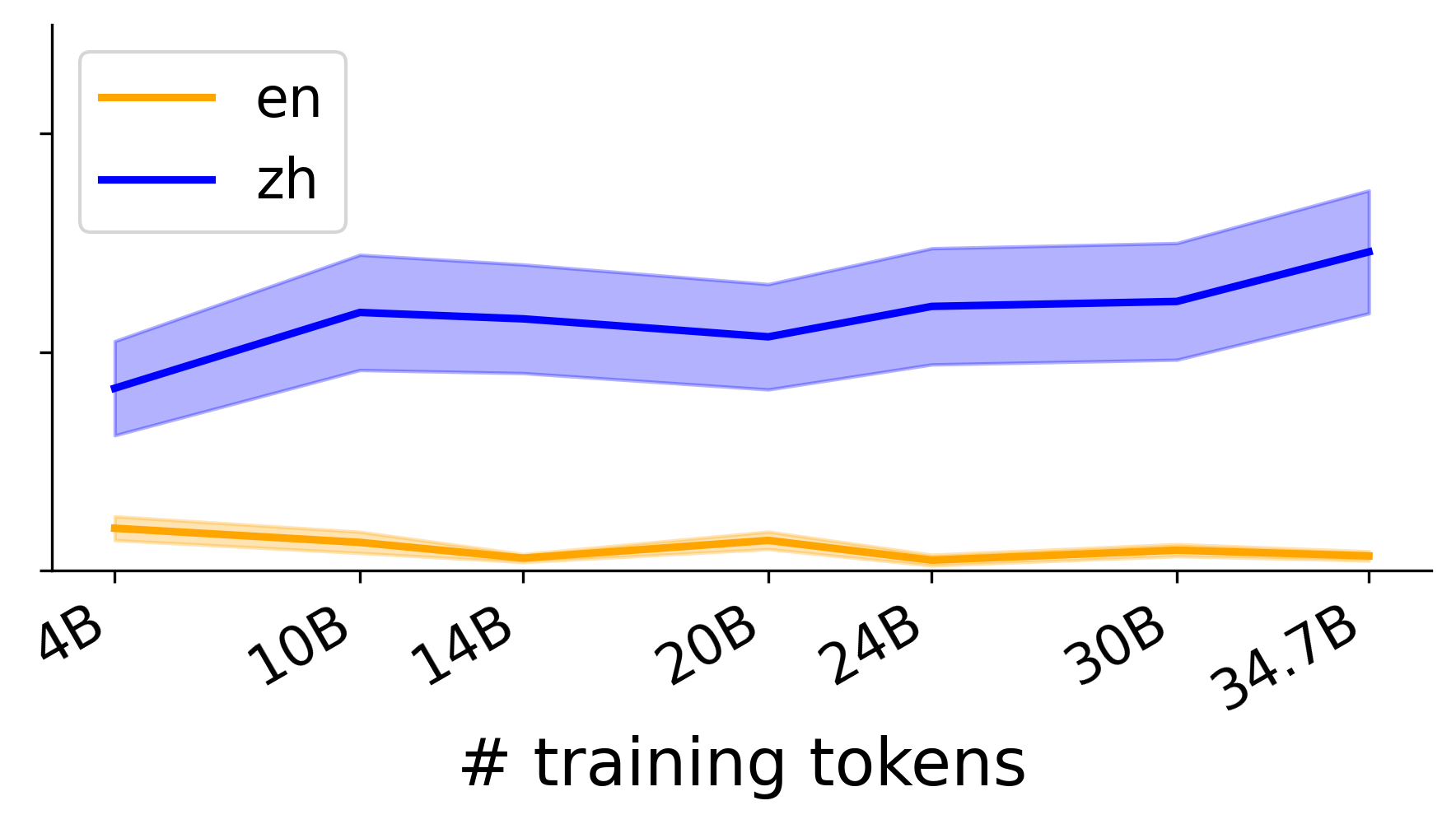}\\
    \small (d) Layer 31
  \end{minipage}\hfill
  \begin{minipage}{0.2\textwidth}
    \centering
    \includegraphics[width=\linewidth]{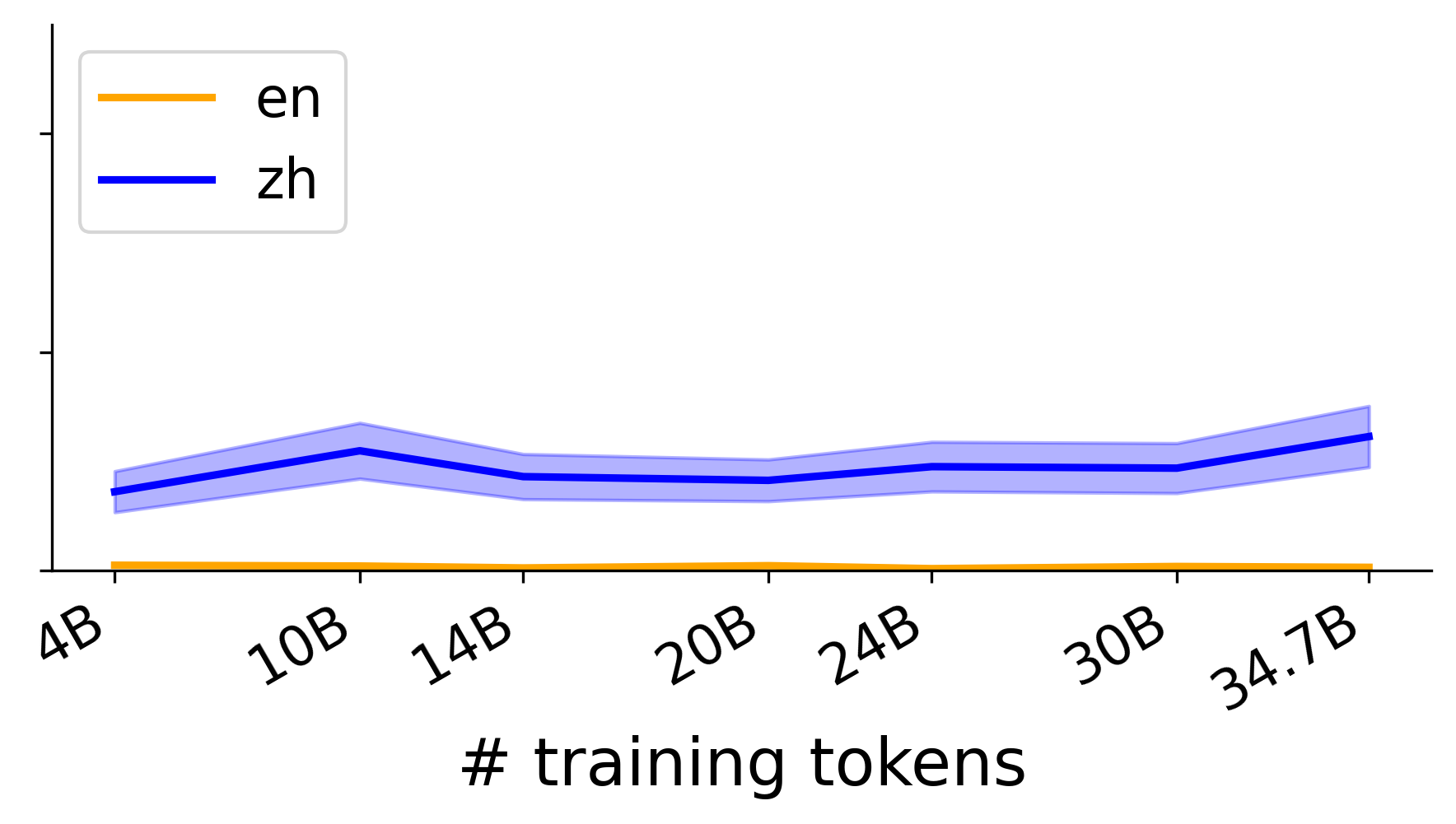}\\
    \small (e) Layer 32
  \end{minipage}
  \caption{Language probabilities of the \textsc{Parallel Only} (7B) model on the Chinese Cloze task across layers and training checkpoints.}
  \label{fig:chinese-cloze-parallel}
\end{figure*}

\begin{figure*}[h!]
  \centering
  \begin{minipage}{0.2\textwidth}
    \centering
    \includegraphics[width=\linewidth]{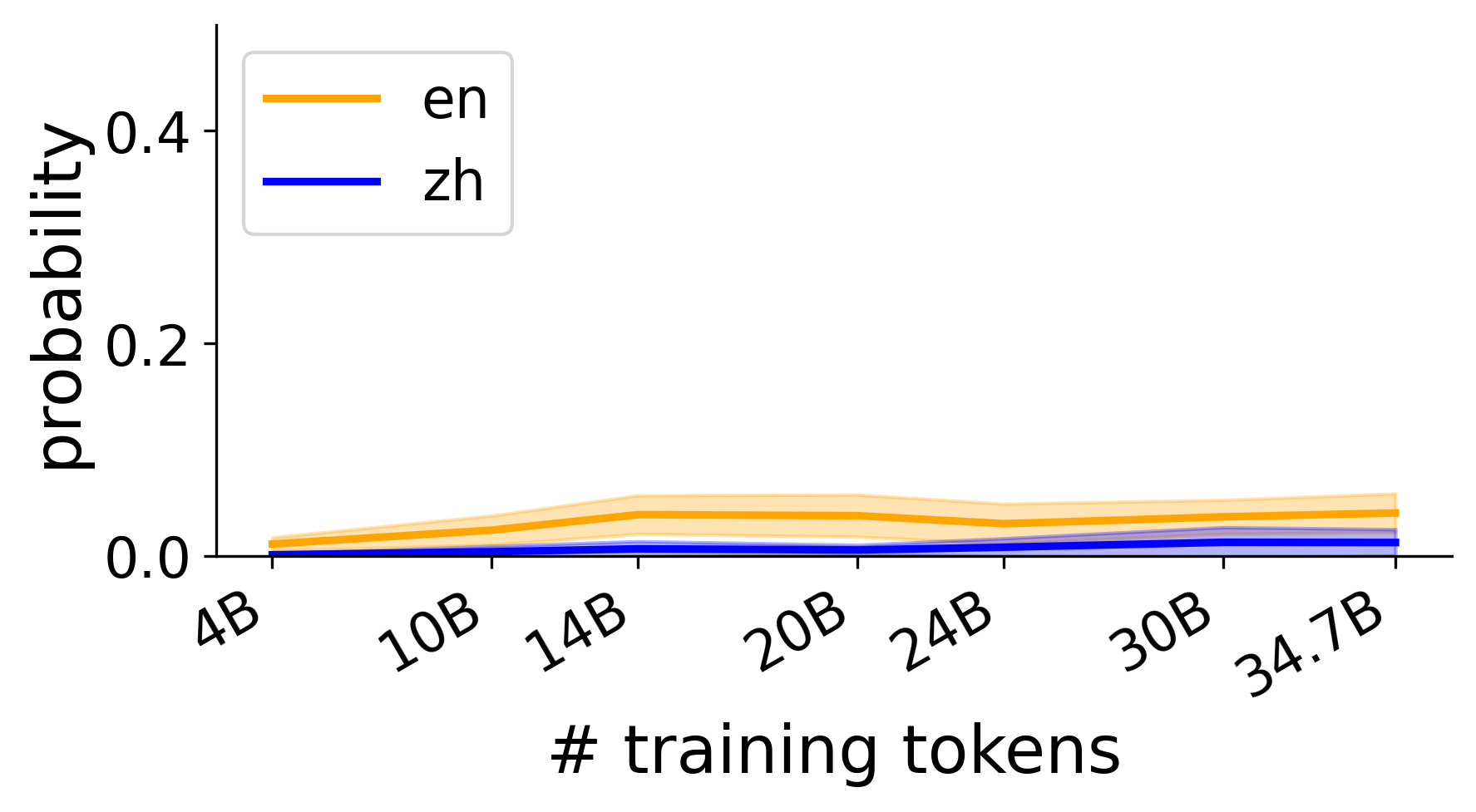}\\
    \small (a) Layer 20
  \end{minipage}\hfill
  \begin{minipage}{0.2\textwidth}
    \centering
    \includegraphics[width=\linewidth]{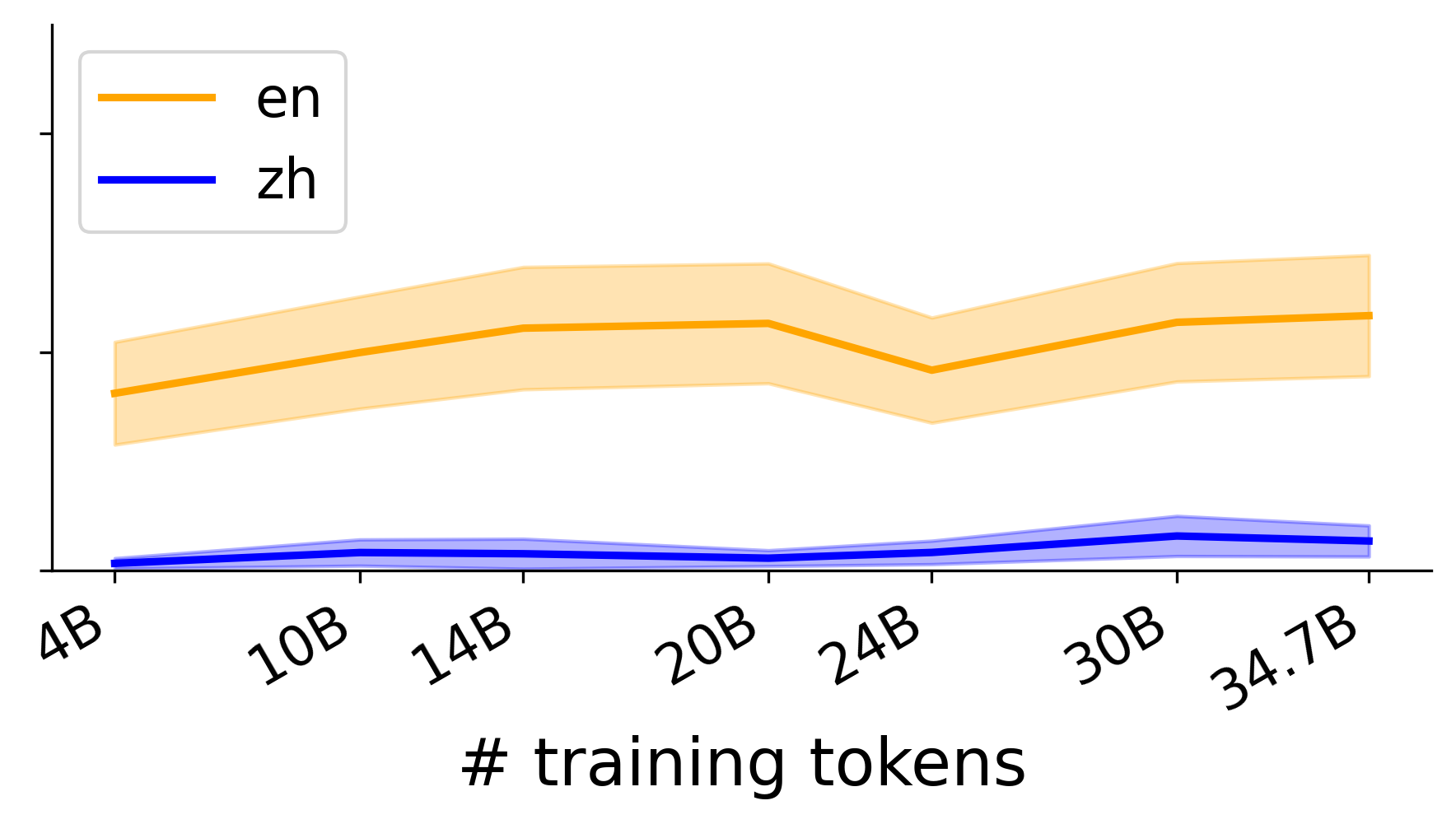}\\
    \small (b) Layer 24
  \end{minipage}\hfill
  \begin{minipage}{0.2\textwidth}
    \centering
    \includegraphics[width=\linewidth]{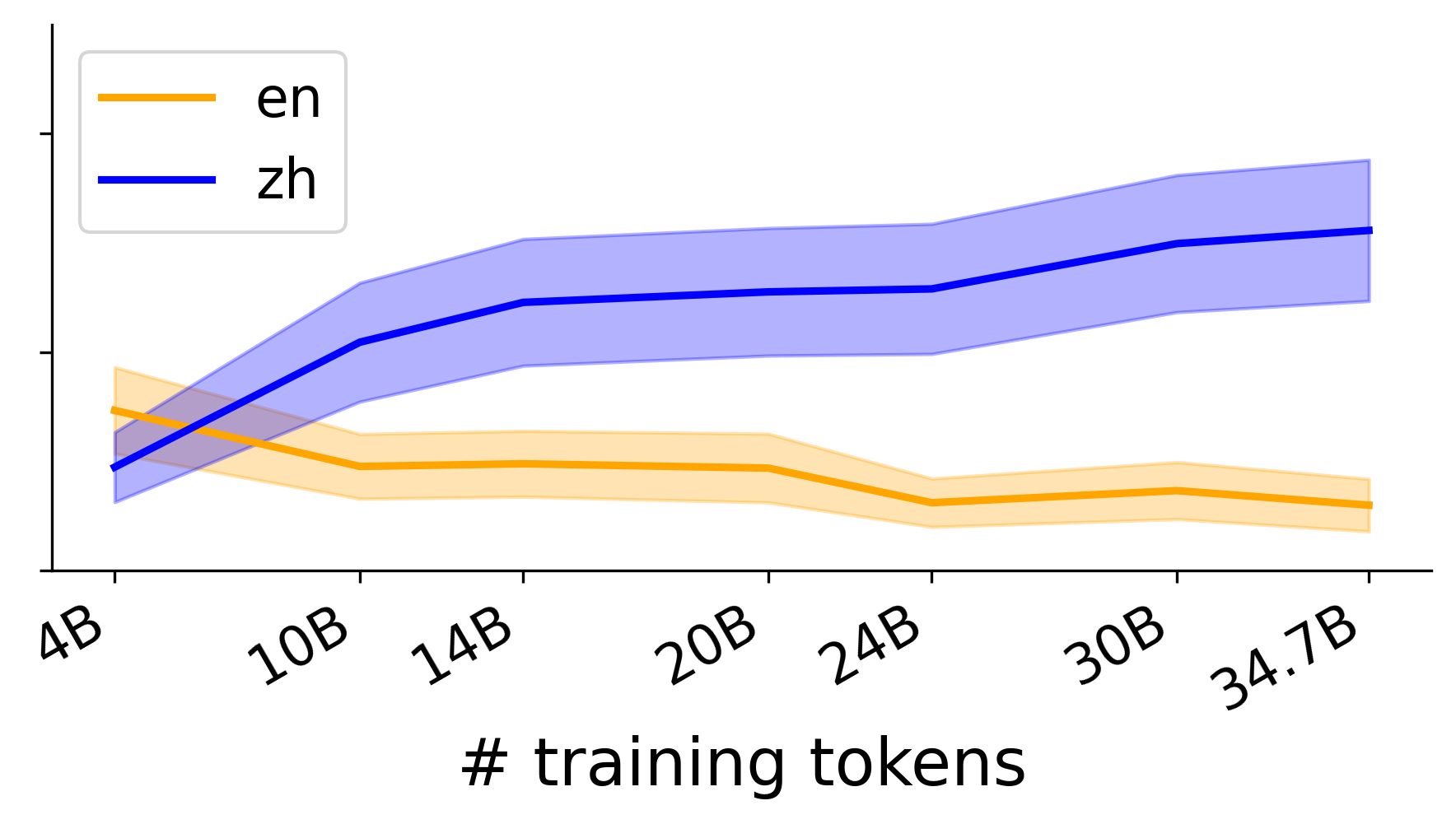}\\
    \small (c) Layer 30
  \end{minipage}\hfill
  \begin{minipage}{0.2\textwidth}
    \centering
    \includegraphics[width=\linewidth]{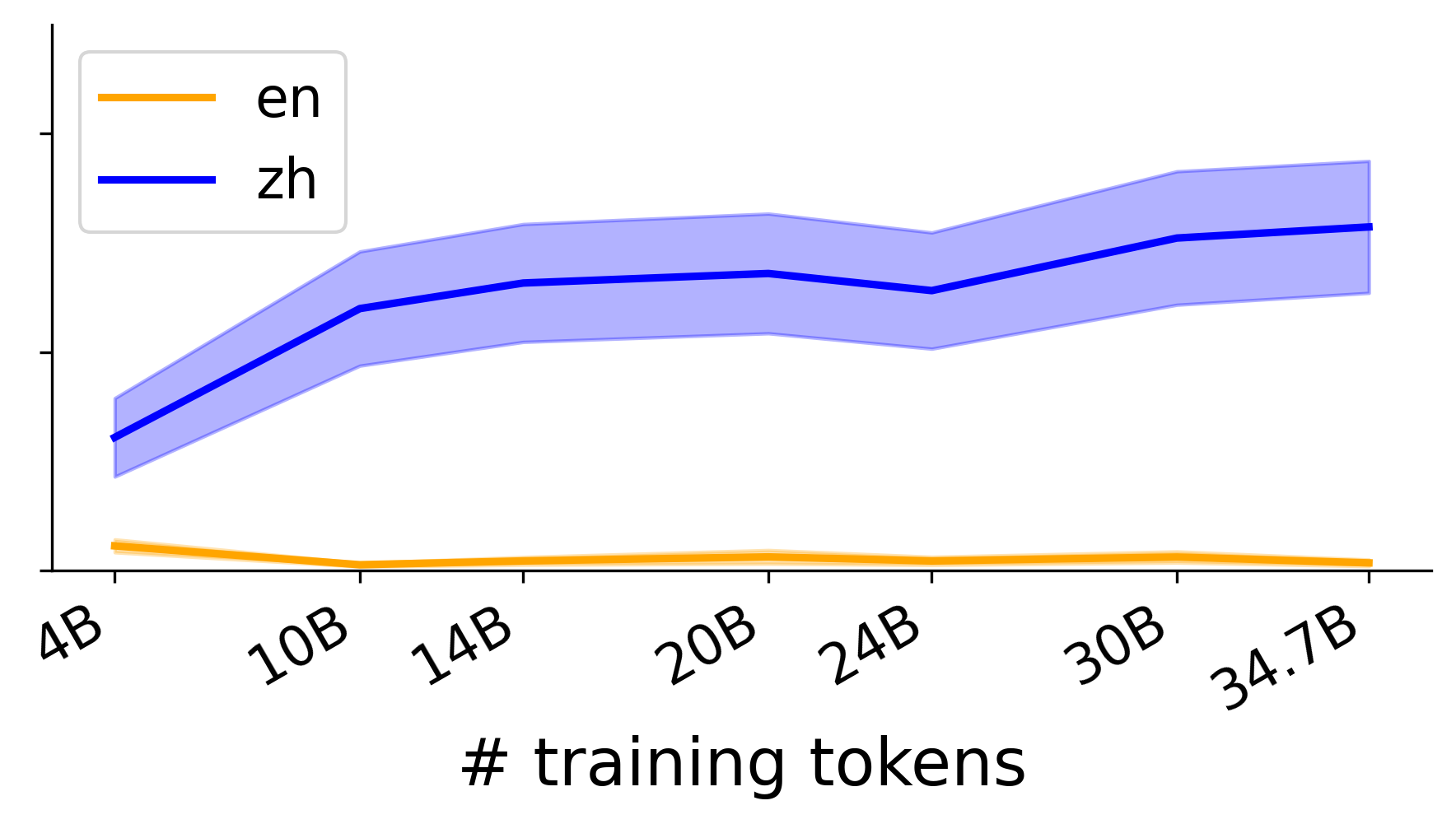}\\
    \small (d) Layer 31
  \end{minipage}\hfill
  \begin{minipage}{0.2\textwidth}
    \centering
    \includegraphics[width=\linewidth]{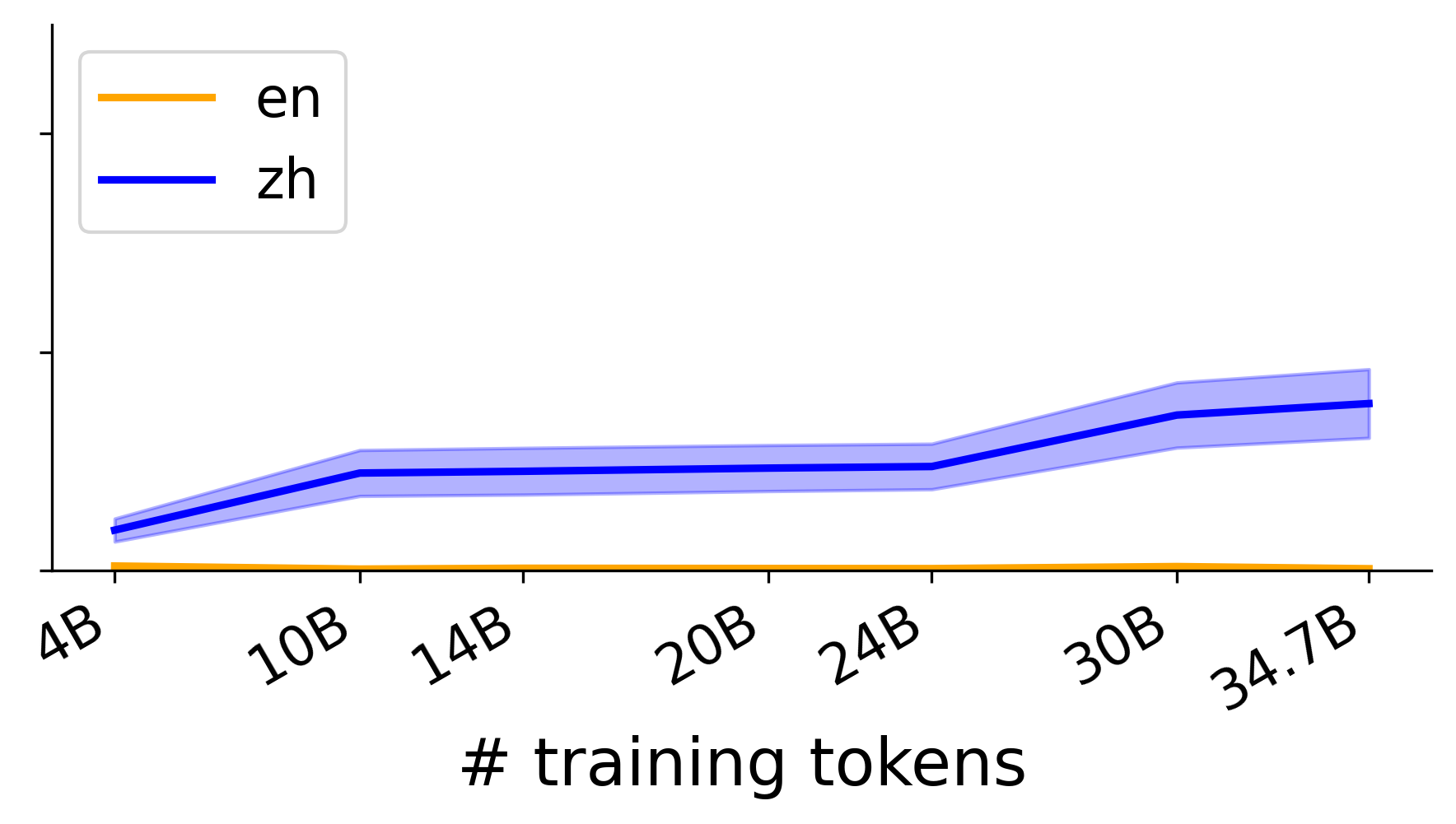}\\
    \small (e) Layer 32
  \end{minipage}
  \caption{Language probabilities of the \textsc{Multilingual} (7B) model on the Chinese Cloze task across layers and training checkpoints.}
  \label{fig:chinese-cloze-multilingual}
\end{figure*}

\subsection{Logit Lens}

Recent work by \citet{wendler-etal-2024-llamas} examined latent language probabilities through the logit lens and demonstrated that multilingual LLMs exhibit a distinctive depth-wise progression at the final training checkpoint: early layers encode minimal language-specific signals, intermediate layers prioritize English representations, and deeper layers shift toward the target language. However, their analysis is limited to a single fully trained model, leaving open questions about how language-specific representations develop over the course of training and how different continual pretraining strategies influence this process.

Figures~\ref{fig:chinese-cloze-parallel} and~\ref{fig:chinese-cloze-multilingual} extend this analysis by tracking language probabilities across training checkpoints for the Chinese Cloze task, comparing the \textsc{Parallel Only} (7B) and \textsc{Multilingual} (7B) models. Because layers below 20 do not exhibit meaningful probabilities for either Chinese or English throughout training, we focus on layers 20 and above, where non-trivial language signals first emerge. Both models reveal a consistent depth-dependent pattern: layer 20 shows weak and unstable English-leaning behavior; intermediate layers (e.g., layer 24) display a pronounced English-dominant phase across training checkpoints; and deeper layers increasingly transition toward Chinese dominance.

Despite these shared patterns, the temporal dynamics of the transition differ between the two models, particularly around layer 30. Figure~\ref{fig:chinese-cloze-parallel} shows that the \textsc{Parallel Only} model undergoes a more gradual shift, with Chinese probability surpassing English only in the later stages of training. In contrast, Figure~\ref{fig:chinese-cloze-multilingual} shows that the \textsc{Multilingual} model exhibits an earlier and more abrupt transition, characterized by a sharp increase in Chinese probabilities and a corresponding decrease in English probability across checkpoints.

English probabilities in the \textsc{Multilingual} model are also generally lower than those in the \textsc{Parallel Only} model, especially in layers 30 and 31. Together, these observations suggest that the \textsc{Parallel Only} model ``thinks'' more in English than the \textsc{Multilingual} model, a behavior also observed in strong multilingual models such as Llama 2 \cite{wendler-etal-2024-llamas}. This may indicate stronger cross-lingual transfer in the \textsc{Parallel Only} model, which could help explain its superior performance across tasks.

\section{Conclusion}
Our study investigates data choices for extending an English-only LLM, OLMo 2, to new languages and demonstrates that, under a fixed token budget, training with only parallel data yields the strongest performance. We further validate this finding on a 7B model continually pretrained on 34.7B tokens, where we compare parallel-only continual pretraining with the more conventional multilingual data approach and observe consistent trends. Through ablation experiments and Logit Lens analyses, we provide evidence that the sentence-level alignment inherent in parallel data plays a key role in its effectiveness, likely by facilitating stronger cross-lingual transfer.

Building on these insights, we introduce \textsc{OpenSeal}, the first fully open Southeast Asian LLM. Our model is fully transparent, enabling careful inspection for security-sensitive applications and promoting AI sovereignty. By releasing models that are entirely open, we aim to support future in-depth research on multilinguality, bias, and other related properties of LLMs.

Despite being trained with a relatively modest computational budget, \textsc{OpenSeal} matches or, in some cases, surpasses existing Southeast Asian LLMs that are derived from high-performing multilingual foundations and trained on substantially larger multilingual corpora. Overall, our results suggest that continually pretraining an English LLM with parallel data offers a good, fast, and cheap pathway to building high-quality, region-focused language models.

\section*{Limitations}
This work focuses less on building the highest-performing LLM and more on empirically investigating and analyzing the effectiveness of parallel data for adapting LLMs to new languages, using Southeast Asian languages as a case study. Our study focuses on the continual pretraining stage and does not include instruction tuning, which could otherwise confound the results. We leave the investigation of instruction tuning to future work. Due to computational resource limitations, our experiments are conducted only at the 1B and 7B parameter scales. We believe that our work poses no immediate risk to society or to any individuals or organizations; however, we recommend caution when using our models, as they have not undergone any safety or value alignment.

\section*{Acknowledgments}

This research is supported by the National Research Foundation Singapore under its AI Singapore Programme (Award Number: AISG3-RP-2022-030). We would like to acknowledge that computational work involved in this research work is partially supported by NUS IT’s Research Computing group with grant number NUSREC-HPC-00001. We would also like to thank Raymond Ng, Peerat Limkonchotiwat, Jian Gang Ngui, and William Tjhi for helpful discussions on SEA-LION. 

\bibliography{custom}

@article{Wolpert1997,
  author = {Wolpert, D.H. and Macready, W.G.},
  doi = {10.1109/4235.585893},
  journal = {IEEE Transactions on Evolutionary Computation},
  month = {April},
  number = 1,
  pages = {67-82},
  title = {No free lunch theorems for optimization},
  volume = 1,
  year = 1997
}

@misc{sengupta2023jaisjaischatarabiccentricfoundation,
      title={Jais and {Jais}-chat: Arabic-Centric Foundation and Instruction-Tuned Open Generative Large Language Models}, 
      author={Neha Sengupta and Sunil Kumar Sahu and Bokang Jia and Satheesh Katipomu and Haonan Li and Fajri Koto and William Marshall and Gurpreet Gosal and Cynthia Liu and Zhiming Chen and Osama Mohammed Afzal and Samta Kamboj and Onkar Pandit and Rahul Pal and Lalit Pradhan and Zain Muhammad Mujahid and Massa Baali and Xudong Han and Sondos Mahmoud Bsharat and Alham Fikri Aji and Zhiqiang Shen and Zhengzhong Liu and Natalia Vassilieva and Joel Hestness and Andy Hock and Andrew Feldman and Jonathan Lee and Andrew Jackson and Hector Xuguang Ren and Preslav Nakov and Timothy Baldwin and Eric Xing},
      year={2023},
      eprint={2308.16149},
      archivePrefix={arXiv},
      primaryClass={cs.CL},
      url={https://arxiv.org/abs/2308.16149}, 
}

@inproceedings{qorib-etal-2025-just,
    title = "Just Go Parallel: Improving the Multilingual Capabilities of Large Language Models",
    author = "Qorib, Muhammad Reza  and
      Li, Junyi  and
      Ng, Hwee Tou",
    editor = "Che, Wanxiang  and
      Nabende, Joyce  and
      Shutova, Ekaterina  and
      Pilehvar, Mohammad Taher",
    booktitle = "Proceedings of ACL",
    month = jul,
    year = "2025",
    IGNOREaddress = "Vienna, Austria",
    IGNOREpublisher = "Association for Computational Linguistics",
    url = "https://aclanthology.org/2025.acl-long.1602/",
    doi = "10.18653/v1/2025.acl-long.1602",
    pages = "33411--33424",
    ISBN = "979-8-89176-251-0",
}

@inproceedings{
zhang2025persistent,
title={Persistent Pre-training Poisoning of {LLM}s},
author={Yiming Zhang and Javier Rando and Ivan Evtimov and Jianfeng Chi and Eric Michael Smith and Nicholas Carlini and Florian Tram{\`e}r and Daphne Ippolito},
booktitle={Proceedings of ICLR},
year={2025},
url={https://openreview.net/forum?id=eiqrnVaeIw}
}

@inproceedings{ConneauL19,
  author       = {Alexis Conneau and
                  Guillaume Lample},
  title        = {Cross-lingual Language Model Pretraining},
  booktitle    = {Proceedings of NeurIPS},
  year         = {2019}
}

@inproceedings{ouyang-etal-2021-ernie,
    title = "{ERNIE}-{M}: Enhanced Multilingual Representation by Aligning Cross-lingual Semantics with Monolingual Corpora",
    author = "Ouyang, Xuan  and
      Wang, Shuohuan  and
      Pang, Chao  and
      Sun, Yu  and
      Tian, Hao  and
      Wu, Hua  and
      Wang, Haifeng",
    booktitle = "Proceedings of EMNLP",
    year = "2021",
    url = "https://aclanthology.org/2021.emnlp-main.3/",
    doi = "10.18653/v1/2021.emnlp-main.3",
    pages = "27--38",
}

@article{liu-etal-2020-multilingual-denoising,
    title = "Multilingual Denoising Pre-training for Neural Machine Translation",
    author = "Liu, Yinhan  and
      Gu, Jiatao  and
      Goyal, Naman  and
      Li, Xian  and
      Edunov, Sergey  and
      Ghazvininejad, Marjan  and
      Lewis, Mike  and
      Zettlemoyer, Luke",
    journal = "TACL",
    volume = "8",
    year = "2020",
    url = "https://aclanthology.org/2020.tacl-1.47/",
    doi = "10.1162/tacl_a_00343",
    pages = "726--742",
}

@inproceedings{chi-etal-2021-mt6,
    title = "m{T}6: Multilingual Pretrained Text-to-Text Transformer with Translation Pairs",
    author = "Chi, Zewen  and
      Dong, Li  and
      Ma, Shuming  and
      Huang, Shaohan  and
      Singhal, Saksham  and
      Mao, Xian-Ling  and
      Huang, Heyan  and
      Song, Xia  and
      Wei, Furu",
    booktitle = "Proceedings of EMNLP",
    year = "2021",
    url = "https://aclanthology.org/2021.emnlp-main.125/",
    doi = "10.18653/v1/2021.emnlp-main.125",
    pages = "1671--1683",
}

@article{bloom,
  author       = {Teven Le Scao and
                  Angela Fan and
                  Christopher Akiki and
                  Ellie Pavlick and
                  Suzana Ilic and
                  Daniel Hesslow and
                  Roman Castagn{\'{e}} and
                  Alexandra Sasha Luccioni and
                  Fran{\c{c}}ois Yvon and
                  Matthias Gall{\'{e}} and
                  Jonathan Tow and
                  Alexander M. Rush and
                  Stella Biderman and
                  Albert Webson and
                  Pawan Sasanka Ammanamanchi and
                  Thomas Wang and
                  Beno{\^{\i}}t Sagot and
                  Niklas Muennighoff and
                  Albert Villanova del Moral and
                  Olatunji Ruwase and
                  Rachel Bawden and
                  Stas Bekman and
                  Angelina McMillan{-}Major and
                  Iz Beltagy and
                  Huu Nguyen and
                  Lucile Saulnier and
                  Samson Tan and
                  Pedro Ortiz Suarez and
                  Victor Sanh and
                  Hugo Lauren{\c{c}}on and
                  Yacine Jernite and
                  Julien Launay and
                  Margaret Mitchell and
                  Colin Raffel and
                  Aaron Gokaslan and
                  Adi Simhi and
                  Aitor Soroa and
                  Alham Fikri Aji and
                  Amit Alfassy and
                  Anna Rogers and
                  Ariel Kreisberg Nitzav and
                  Canwen Xu and
                  Chenghao Mou and
                  Chris Emezue and
                  Christopher Klamm and
                  Colin Leong and
                  Daniel van Strien and
                  David Ifeoluwa Adelani and
                  et al.},
  title        = "{BLOOM}: A 176{B}-Parameter Open-Access Multilingual Language Model",
  journal={arXiv preprint arXiv:2211.05100},
  year         = {2022}
}

@inproceedings{post-2018-call,
    title = "A Call for Clarity in Reporting {BLEU} Scores",
    author = "Post, Matt",
    editor = "Bojar, Ond{\v{r}}ej  and
      Chatterjee, Rajen  and
      Federmann, Christian  and
      Fishel, Mark  and
      Graham, Yvette  and
      Haddow, Barry  and
      Huck, Matthias  and
      Yepes, Antonio Jimeno  and
      Koehn, Philipp  and
      Monz, Christof  and
      Negri, Matteo  and
      N{\'e}v{\'e}ol, Aur{\'e}lie  and
      Neves, Mariana  and
      Post, Matt  and
      Specia, Lucia  and
      Turchi, Marco  and
      Verspoor, Karin",
    booktitle = "Proceedings of WMT",
    month = oct,
    year = "2018",
    IGNOREaddress = "Brussels, Belgium",
    IGNOREpublisher = "Association for Computational Linguistics",
    url = "https://aclanthology.org/W18-6319/",
    doi = "10.18653/v1/W18-6319",
    pages = "186--191"
}

@inproceedings{conneau-etal-2018-xnli,
    title = "{XNLI}: Evaluating Cross-lingual Sentence Representations",
    author = "Conneau, Alexis  and
      Rinott, Ruty  and
      Lample, Guillaume  and
      Williams, Adina  and
      Bowman, Samuel  and
      Schwenk, Holger  and
      Stoyanov, Veselin",
    editor = "Riloff, Ellen  and
      Chiang, David  and
      Hockenmaier, Julia  and
      Tsujii, Jun{'}ichi",
    booktitle = "Proceedings of EMNLP",
    month = oct # "-" # nov,
    year = "2018",
    IGNOREaddress = "Brussels, Belgium",
    url = "https://aclanthology.org/D18-1269/",
    doi = "10.18653/v1/D18-1269",
    pages = "2475--2485"
}

@inproceedings{ponti-etal-2020-xcopa,
    title = "{XCOPA}: A Multilingual Dataset for Causal Commonsense Reasoning",
    author = "Ponti, Edoardo Maria  and
      Glava{\v{s}}, Goran  and
      Majewska, Olga  and
      Liu, Qianchu  and
      Vuli{\'c}, Ivan  and
      Korhonen, Anna",
    editor = "Webber, Bonnie  and
      Cohn, Trevor  and
      He, Yulan  and
      Liu, Yang",
    booktitle = "Proceedings of EMNLP",
    month = nov,
    year = "2020",
    IGNOREaddress = "Online",
    IGNOREpublisher = "Association for Computational Linguistics",
    url = "https://aclanthology.org/2020.emnlp-main.185/",
    doi = "10.18653/v1/2020.emnlp-main.185",
    pages = "2362--2376"
}

@inproceedings{yang-etal-2019-paws,
    title = "{PAWS}-{X}: A Cross-lingual Adversarial Dataset for Paraphrase Identification",
    author = "Yang, Yinfei  and
      Zhang, Yuan  and
      Tar, Chris  and
      Baldridge, Jason",
    editor = "Inui, Kentaro  and
      Jiang, Jing  and
      Ng, Vincent  and
      Wan, Xiaojun",
    booktitle = "Proceedings of EMNLP-IJCNLP",
    month = nov,
    year = "2019",
    IGNOREaddress = "Hong Kong, China",
    IGNOREpublisher = "Association for Computational Linguistics",
    url = "https://aclanthology.org/D19-1382/",
    doi = "10.18653/v1/D19-1382",
    pages = "3687--3692"
}

@misc{apertus2025apertusdemocratizingopencompliant,
      title={Apertus: Democratizing Open and Compliant {LLMs} for Global Language Environments}, 
      author={Alejandro Hernández-Cano and Alexander Hägele and Allen Hao Huang and Angelika Romanou and Antoni-Joan Solergibert and Barna Pasztor and Bettina Messmer and Dhia Garbaya and Eduard Frank Ďurech and Ido Hakimi and Juan García Giraldo and Mete Ismayilzada and Negar Foroutan and Skander Moalla and Tiancheng Chen and Vinko Sabolčec and Yixuan Xu and Michael Aerni and Badr AlKhamissi and Inés Altemir Mariñas and Mohammad Hossein Amani and Matin Ansaripour and Ilia Badanin and Harold Benoit and Emanuela Boros and Nicholas Browning and Fabian Bösch and Maximilian Böther and Niklas Canova and Camille Challier and Clement Charmillot and Jonathan Coles and Jan Deriu and Arnout Devos and Lukas Drescher and Daniil Dzenhaliou and Maud Ehrmann and Dongyang Fan and Simin Fan and Silin Gao and Miguel Gila and María Grandury and Diba Hashemi and Alexander Hoyle and Jiaming Jiang and Mark Klein and Andrei Kucharavy and Anastasiia Kucherenko and Frederike Lübeck and Roman Machacek and Theofilos Manitaras and Andreas Marfurt and Kyle Matoba and Simon Matrenok and Henrique Mendonça and Fawzi Roberto Mohamed and Syrielle Montariol and Luca Mouchel and Sven Najem-Meyer and Jingwei Ni and Gennaro Oliva and Matteo Pagliardini and Elia Palme and Andrei Panferov and Léo Paoletti and Marco Passerini and Ivan Pavlov and Auguste Poiroux and Kaustubh Ponkshe and Nathan Ranchin and Javi Rando and Mathieu Sauser and Jakhongir Saydaliev and Muhammad Ali Sayfiddinov and Marian Schneider and Stefano Schuppli and Marco Scialanga and Andrei Semenov and Kumar Shridhar and Raghav Singhal and Anna Sotnikova and Alexander Sternfeld and Ayush Kumar Tarun and Paul Teiletche and Jannis Vamvas and Xiaozhe Yao and Hao Zhao and Alexander Ilic and Ana Klimovic and Andreas Krause and Caglar Gulcehre and David Rosenthal and Elliott Ash and Florian Tramèr and Joost VandeVondele and Livio Veraldi and Martin Rajman and Thomas Schulthess and Torsten Hoefler and Antoine Bosselut and Martin Jaggi and Imanol Schlag},
      year={2025},
      eprint={2509.14233},
      archivePrefix={arXiv},
      primaryClass={cs.CL},
      url={https://arxiv.org/abs/2509.14233}, 
}

@inproceedings{
fujii2024continual,
title={Continual Pre-Training for Cross-Lingual {LLM} Adaptation: Enhancing {Japanese} Language Capabilities},
author={Kazuki Fujii and Taishi Nakamura and Mengsay Loem and Hiroki Iida and Masanari Ohi and Kakeru Hattori and Hirai Shota and Sakae Mizuki and Rio Yokota and Naoaki Okazaki},
booktitle={Proceedings of COLM},
year={2024},
url={https://openreview.net/forum?id=TQdd1VhWbe}
}

@inproceedings{
walsh2025,
title={2 {OLM}o 2 Furious},
author={Evan Pete Walsh and Luca Soldaini and Dirk Groeneveld and Kyle Lo and Shane Arora and Akshita Bhagia and Yuling Gu and Shengyi Huang and Matt Jordan and Nathan Lambert and Dustin Schwenk and Oyvind Tafjord and Taira Anderson and David Atkinson and Faeze Brahman and Christopher Clark and Pradeep Dasigi and Nouha Dziri and Allyson Ettinger and Michal Guerquin and David Heineman and Hamish Ivison and Pang Wei Koh and Jiacheng Liu and Saumya Malik and William Merrill and Lester James Validad Miranda and Jacob Morrison and Tyler Murray and Crystal Nam and Jake Poznanski and Valentina Pyatkin and Aman Rangapur and Michael Schmitz and Sam Skjonsberg and David Wadden and Christopher Wilhelm and Michael Wilson and Luke Zettlemoyer and Ali Farhadi and Noah A. Smith and Hannaneh Hajishirzi},
booktitle={Proceedings of COLM},
year={2025},
url={https://openreview.net/forum?id=2ezugTT9kU}
}

@inproceedings{zheng-etal-2024-breaking,
    title = "Breaking Language Barriers: Cross-Lingual Continual Pre-Training at Scale",
    author = "Zheng, Wenzhen  and
      Pan, Wenbo  and
      Xu, Xu  and
      Qin, Libo  and
      Yue, Li  and
      Zhou, Ming",
    editor = "Al-Onaizan, Yaser  and
      Bansal, Mohit  and
      Chen, Yun-Nung",
    booktitle = "Proceedings of EMNLP",
    month = nov,
    year = "2024",
    IGNOREaddress = "Miami, Florida, USA",
    IGNOREpublisher = "Association for Computational Linguistics",
    url = "https://aclanthology.org/2024.emnlp-main.441/",
    doi = "10.18653/v1/2024.emnlp-main.441",
    pages = "7725--7738",
}

@inproceedings{ranaldi-etal-2024-empowering,
    title = "Empowering cross-lingual abilities of instruction-tuned large language models by translation-following demonstrations",
    author = "Ranaldi, Leonardo  and
      Pucci, Giulia  and
      Freitas, Andre",
    editor = "Ku, Lun-Wei  and
      Martins, Andre  and
      Srikumar, Vivek",
    booktitle = "Findings of ACL",
    month = aug,
    year = "2024",
    IGNOREaddress = "Bangkok, Thailand",
    IGNOREpublisher = "Association for Computational Linguistics",
    url = "https://aclanthology.org/2024.findings-acl.473/",
    doi = "10.18653/v1/2024.findings-acl.473",
    pages = "7961--7973",
}

@article{llama2,
  author       = {Hugo Touvron and
                  Louis Martin and
                  Kevin Stone and
                  Peter Albert and
                  Amjad Almahairi and
                  Yasmine Babaei and
                  Nikolay Bashlykov and
                  Soumya Batra and
                  Prajjwal Bhargava and
                  Shruti Bhosale and
                  Dan Bikel and
                  Lukas Blecher and
                  Cristian Canton{-}Ferrer and
                  Moya Chen and
                  Guillem Cucurull and
                  David Esiobu and
                  Jude Fernandes and
                  Jeremy Fu and
                  Wenyin Fu and
                  Brian Fuller and
                  Cynthia Gao and
                  Vedanuj Goswami and
                  Naman Goyal and
                  Anthony Hartshorn and
                  Saghar Hosseini and
                  Rui Hou and
                  Hakan Inan and
                  Marcin Kardas and
                  Viktor Kerkez and
                  Madian Khabsa and
                  Isabel Kloumann and
                  Artem Korenev and
                  Punit Singh Koura and
                  Marie{-}Anne Lachaux and
                  Thibaut Lavril and
                  Jenya Lee and
                  Diana Liskovich and
                  Yinghai Lu and
                  Yuning Mao and
                  Xavier Martinet and
                  Todor Mihaylov and
                  Pushkar Mishra and
                  Igor Molybog and
                  Yixin Nie and
                  Andrew Poulton and
                  Jeremy Reizenstein and
                  Rashi Rungta and
                  Kalyan Saladi and
                  Alan Schelten and
                  Ruan Silva and
                  Eric Michael Smith and
                  Ranjan Subramanian and
                  Xiaoqing Ellen Tan and
                  Binh Tang and
                  Ross Taylor and
                  Adina Williams and
                  Jian Xiang Kuan and
                  Puxin Xu and
                  Zheng Yan and
                  Iliyan Zarov and
                  Yuchen Zhang and
                  Angela Fan and
                  Melanie Kambadur and
                  Sharan Narang and
                  Aur{\'{e}}lien Rodriguez and
                  Robert Stojnic and
                  Sergey Edunov and
                  Thomas Scialom},
  title        = {Llama 2: {O}pen Foundation and Fine-Tuned Chat Models},
  journal={arXiv preprint arXiv:2307.09288},
  year         = {2023}
}

@misc{ng2025sealionsoutheastasianlanguages,
Title = {SEA-LION: Southeast {Asian} Languages in One Network},
Author = {Raymond Ng and Thanh Ngan Nguyen and Yuli Huang and Ngee Chia Tai and Wai Yi Leong and Wei Qi Leong and Xianbin Yong and Jian Gang Ngui and Yosephine Susanto and Nicholas Cheng and Hamsawardhini Rengarajan and Peerat Limkonchotiwat and Adithya Venkatadri Hulagadri and Kok Wai Teng and Yeo Yeow Tong and Bryan Siow and Wei Yi Teo and Wayne Lau and Choon Meng Tan and Brandon Ong and Zhi Hao Ong and Jann Railey Montalan and Adwin Chan and Sajeban Antonyrex and Ren Lee and Esther Choa and David Ong Tat-Wee and Bing Jie Darius Liu and William Chandra Tjhi and Erik Cambria and Leslie Teo},
Year = {2025},
Eprint = {2504.05747},
archivePrefix={arXiv},
primaryClass={cs.CL},
url={https://arxiv.org/abs/2504.05747}
}

@misc{du2024chinesetinyllmpretraining,
      title={Chinese {Tiny} {LLM}: Pretraining a {Chinese}-Centric Large Language Model}, 
      author={Xinrun Du and Zhouliang Yu and Songyang Gao and Ding Pan and Yuyang Cheng and Ziyang Ma and Ruibin Yuan and Xingwei Qu and Jiaheng Liu and Tianyu Zheng and Xinchen Luo and Guorui Zhou and Wenhu Chen and Ge Zhang},
      year={2024},
      eprint={2404.04167},
      archivePrefix={arXiv},
      primaryClass={cs.CL},
      url={https://arxiv.org/abs/2404.04167}, 
}

@inproceedings{
xu2024a,
title={A Paradigm Shift in Machine Translation: Boosting Translation Performance of Large Language Models},
author={Haoran Xu and Young Jin Kim and Amr Sharaf and Hany Hassan Awadalla},
booktitle={Proceedings of ICLR},
year={2024},
url={https://openreview.net/forum?id=farT6XXntP}
}

@misc{ibrahim2024simplescalablestrategiescontinually,
      title={Simple and Scalable Strategies to Continually Pre-train Large Language Models}, 
      author={Adam Ibrahim and Benjamin Thérien and Kshitij Gupta and Mats L. Richter and Quentin Anthony and Timothée Lesort and Eugene Belilovsky and Irina Rish},
      year={2024},
      eprint={2403.08763},
      archivePrefix={arXiv},
      primaryClass={cs.LG},
      url={https://arxiv.org/abs/2403.08763}, 
}

@inproceedings{nguyen-etal-2024-seallms,
    title = "{S}ea{LLM}s - Large Language Models for {S}outheast {A}sia",
    author = "Nguyen, Xuan-Phi  and
      Zhang, Wenxuan  and
      Li, Xin  and
      Aljunied, Mahani  and
      Hu, Zhiqiang  and
      Shen, Chenhui  and
      Chia, Yew Ken  and
      Li, Xingxuan  and
      Wang, Jianyu  and
      Tan, Qingyu  and
      Cheng, Liying  and
      Chen, Guanzheng  and
      Deng, Yue  and
      Yang, Sen  and
      Liu, Chaoqun  and
      Zhang, Hang  and
      Bing, Lidong",
    editor = "Cao, Yixin  and
      Feng, Yang  and
      Xiong, Deyi",
    booktitle = "Proceedings of ACL",
    month = aug,
    year = "2024",
    IGNOREaddress = "Bangkok, Thailand",
    IGNOREpublisher = "Association for Computational Linguistics",
    url = "https://aclanthology.org/2024.acl-demos.28/",
    doi = "10.18653/v1/2024.acl-demos.28",
    pages = "294--304",
}

@inproceedings{zhang-etal-2025-seallms,
    title = "{S}ea{LLM}s 3: Open Foundation and Chat Multilingual Large Language Models for {S}outheast {A}sian Languages",
    author = "Zhang, Wenxuan  and
      Chan, Hou Pong  and
      Zhao, Yiran  and
      Aljunied, Mahani  and
      Wang, Jianyu  and
      Liu, Chaoqun  and
      Deng, Yue  and
      Hu, Zhiqiang  and
      Xu, Weiwen  and
      Chia, Yew Ken  and
      Li, Xin  and
      Bing, Lidong",
    editor = "Dziri, Nouha  and
      Ren, Sean (Xiang)  and
      Diao, Shizhe",
    booktitle = "Proceedings of ACL",
    month = apr,
    year = "2025",
    IGNOREaddress = "Albuquerque, New Mexico",
    IGNOREpublisher = "Association for Computational Linguistics",
    url = "https://aclanthology.org/2025.naacl-demo.10/",
    doi = "10.18653/v1/2025.naacl-demo.10",
    pages = "96--105",
    ISBN = "979-8-89176-191-9",
}

@inproceedings{wu-etal-2024-llama,
    title = "{LL}a{MA} Pro: Progressive {LL}a{MA} with Block Expansion",
    author = "Wu, Chengyue  and
      Gan, Yukang  and
      Ge, Yixiao  and
      Lu, Zeyu  and
      Wang, Jiahao  and
      Feng, Ye  and
      Shan, Ying  and
      Luo, Ping",
    editor = "Ku, Lun-Wei  and
      Martins, Andre  and
      Srikumar, Vivek",
    booktitle = "Proceedings of ACL",
    month = aug,
    year = "2024",
    IGNOREaddress = "Bangkok, Thailand",
    IGNOREpublisher = "Association for Computational Linguistics",
    url = "https://aclanthology.org/2024.acl-long.352/",
    doi = "10.18653/v1/2024.acl-long.352",
    pages = "6518--6537",
}

@inproceedings{dou-etal-2024-sailor,
    title = "Sailor: Open Language Models for {S}outh-{E}ast {A}sia",
    author = "Dou, Longxu  and
      Liu, Qian  and
      Zeng, Guangtao  and
      Guo, Jia  and
      Zhou, Jiahui  and
      Mao, Xin  and
      Jin, Ziqi  and
      Lu, Wei  and
      Lin, Min",
    editor = "Hernandez Farias, Delia Irazu  and
      Hope, Tom  and
      Li, Manling",
    booktitle = "Proceedings of EMNLP",
    month = nov,
    year = "2024",
    IGNOREaddress = "Miami, Florida, USA",
    IGNOREpublisher = "Association for Computational Linguistics",
    url = "https://aclanthology.org/2024.emnlp-demo.45/",
    doi = "10.18653/v1/2024.emnlp-demo.45",
    pages = "424--435",
}

@inproceedings{wendler-etal-2024-llamas,
    title = "Do {Llamas} Work in {E}nglish? On the Latent Language of Multilingual Transformers",
    author = "Wendler, Chris  and
      Veselovsky, Veniamin  and
      Monea, Giovanni  and
      West, Robert",
    editor = "Ku, Lun-Wei  and
      Martins, Andre  and
      Srikumar, Vivek",
    booktitle = "Proceedings of ACL",
    month = aug,
    year = "2024",
    IGNOREaddress = "Bangkok, Thailand",
    IGNOREpublisher = "Association for Computational Linguistics",
    url = "https://aclanthology.org/2024.acl-long.820/",
    doi = "10.18653/v1/2024.acl-long.820",
    pages = "15366--15394"
}

@inproceedings{lovenia-etal-2024-seacrowd,
    title = "{SEAC}rowd: A Multilingual Multimodal Data Hub and Benchmark Suite for {S}outheast {A}sian Languages",
    author = {Lovenia, Holy  and
      Mahendra, Rahmad  and
      Akbar, Salsabil Maulana  and
      Miranda, Lester James V.  and
      Santoso, Jennifer  and
      Aco, Elyanah  and
      Fadhilah, Akhdan  and
      Mansurov, Jonibek  and
      Imperial, Joseph Marvin  and
      Kampman, Onno P.  and
      Moniz, Joel Ruben Antony  and
      Habibi, Muhammad Ravi Shulthan  and
      Hudi, Frederikus  and
      Montalan, Railey  and
      Ignatius, Ryan  and
      Lopo, Joanito Agili  and
      Nixon, William  and
      Karlsson, B{\"o}rje F.  and
      Jaya, James  and
      Diandaru, Ryandito  and
      Gao, Yuze  and
      Amadeus, Patrick  and
      Wang, Bin  and
      Cruz, Jan Christian Blaise  and
      Whitehouse, Chenxi  and
      Parmonangan, Ivan Halim  and
      Khelli, Maria  and
      Zhang, Wenyu  and
      Susanto, Lucky  and
      Ryanda, Reynard Adha  and
      Hermawan, Sonny Lazuardi  and
      Velasco, Dan John  and
      Kautsar, Muhammad Dehan Al  and
      Hendria, Willy Fitra  and
      Moslem, Yasmin  and
      Flynn, Noah  and
      Adilazuarda, Muhammad Farid  and
      Li, Haochen  and
      Lee, Johanes  and
      Damanhuri, R.  and
      Sun, Shuo  and
      Qorib, Muhammad Reza  and
      Djanibekov, Amirbek  and
      Leong, Wei Qi  and
      Do, Quyet V.  and
      Muennighoff, Niklas  and
      Pansuwan, Tanrada  and
      Putra, Ilham Firdausi  and
      Xu, Yan  and
      Chia, Tai Ngee  and
      Purwarianti, Ayu  and
      Ruder, Sebastian  and
      Tjhi, William  and
      Limkonchotiwat, Peerat  and
      Aji, Alham Fikri  and
      Keh, Sedrick  and
      Winata, Genta Indra  and
      Zhang, Ruochen  and
      Koto, Fajri  and
      Yong, Zheng-Xin  and
      Cahyawijaya, Samuel},
    editor = "Al-Onaizan, Yaser  and
      Bansal, Mohit  and
      Chen, Yun-Nung",
    booktitle = "Proceedings of EMNLP",
    month = nov,
    year = "2024",
    IGNOREaddress = "Miami, Florida, USA",
    IGNOREpublisher = "Association for Computational Linguistics",
    url = "https://aclanthology.org/2024.emnlp-main.296/",
    doi = "10.18653/v1/2024.emnlp-main.296",
    pages = "5155--5203",
}

@article{10.5555/3648699.3648939,
author = {Chowdhery, Aakanksha and Narang, Sharan and Devlin, Jacob and Bosma, Maarten and Mishra, Gaurav and Roberts, Adam and Barham, Paul and Chung, Hyung Won and Sutton, Charles and Gehrmann, Sebastian and et al.},
title = {{PaLM}: Scaling language modeling with pathways},
year = {2023},
issue_date = {January 2023},
volume = {24},
number = {1},
issn = {1532-4435},
journal = {JMLR},
articleno = {240},
numpages = {113},
}

@inproceedings{briakou-etal-2023-searching,
    title = "Searching for Needles in a Haystack: On the Role of Incidental Bilingualism in {P}a{LM}`s Translation Capability",
    author = "Briakou, Eleftheria  and
      Cherry, Colin  and
      Foster, George",
    booktitle = "Proceedings of ACL",
    year = "2023",
    url = "https://aclanthology.org/2023.acl-long.524/",
    doi = "10.18653/v1/2023.acl-long.524",
    pages = "9432--9452",
}

@inproceedings{papineni-etal-2002-bleu,
    title = "{B}leu: a Method for Automatic Evaluation of Machine Translation",
    author = "Papineni, Kishore  and
      Roukos, Salim  and
      Ward, Todd  and
      Zhu, Wei-Jing",
    editor = "Isabelle, Pierre  and
      Charniak, Eugene  and
      Lin, Dekang",
    booktitle = "Proceedings of ACL",
    month = jul,
    year = "2002",
    IGNOREaddress = "Philadelphia, Pennsylvania, USA",
    IGNOREpublisher = "Association for Computational Linguistics",
    url = "https://aclanthology.org/P02-1040/",
    doi = "10.3115/1073083.1073135",
    pages = "311--318"
}

@misc{nllb2022,
      title={No Language Left Behind: Scaling Human-Centered Machine Translation}, 
      author={Marta R. Costa-jussà and James Cross and Onur Çelebi and Maha Elbayad and Kenneth Heafield and Kevin Heffernan and Elahe Kalbassi and Janice Lam and Daniel Licht and Jean Maillard and Anna Sun and Skyler Wang and Guillaume Wenzek and Al Youngblood and Bapi Akula and Loic Barrault and Gabriel Mejia Gonzalez and Prangthip Hansanti and John Hoffman and Semarley Jarrett and Kaushik Ram Sadagopan and Dirk Rowe and Shannon Spruit and Chau Tran and Pierre Andrews and Necip Fazil Ayan and Shruti Bhosale and Sergey Edunov and Angela Fan and Cynthia Gao and Vedanuj Goswami and Francisco Guzmán and Philipp Koehn and Alexandre Mourachko and Christophe Ropers and Safiyyah Saleem and Holger Schwenk and Jeff Wang},
      year={2022},
      eprint={2207.04672},
      archivePrefix={arXiv},
      primaryClass={cs.CL},
      url={https://arxiv.org/abs/2207.04672}, 
}

@inproceedings{
hu2024minicpm,
title={Mini{CPM}: Unveiling the Potential of Small Language Models with Scalable Training Strategies},
author={Shengding Hu and Yuge Tu and Xu Han and Ganqu Cui and Chaoqun He and Weilin Zhao and Xiang Long and Zhi Zheng and Yewei Fang and Yuxiang Huang and Xinrong Zhang and Zhen Leng Thai and Chongyi Wang and Yuan Yao and Chenyang Zhao and Jie Zhou and Jie Cai and Zhongwu Zhai and Ning Ding and Chao Jia and Guoyang Zeng and dahai li and Zhiyuan Liu and Maosong Sun},
booktitle={Proceedings of COLM},
year={2024},
url={https://openreview.net/forum?id=3X2L2TFr0f}
}

@inproceedings{koehn-2004-statistical,
    title = "Statistical Significance Tests for Machine Translation Evaluation",
    author = "Koehn, Philipp",
    editor = "Lin, Dekang  and
      Wu, Dekai",
    booktitle = "Proceedings of EMNLP",
    month = jul,
    year = "2004",
    IGNOREaddress = "Barcelona, Spain",
    IGNOREpublisher = "Association for Computational Linguistics",
    url = "https://aclanthology.org/W04-3250/",
    pages = "388--395"
}

\appendix

\section{Resources}
\label{sec:resources}
In this section, we report the details of the data sources used in our experiments. We report the statistics of our SEA multilingual data (Table~\ref{tab:sta-sea_pile_v2}), Chinese multilingual data (Table~\ref{tab:map-cc}), and replay data (Table~\ref{tab:olmo2-replay}).

\begin{table}
    \centering
    \begin{tabular}{lr}
    \hline
    \textbf{ISO Code} & \textbf{\# tokens} \\
    \hline
    \texttt{id} & 54.6B\\
    \texttt{km} & 1.7B\\
    \texttt{lo} & 2.1B\\
    \texttt{ms} & 11.8B\\
    \texttt{my} & 0.8B\\
    \texttt{ta} & 46.3B\\
    \texttt{th} & 16.5B\\
    \texttt{tl} & 2.5B\\
    \texttt{vi} & 90.0B\\
    \hline
    \end{tabular}
    \caption{Number of tokens of SEA-PILE-v2, measured using the OLMo 2 tokenizer.}
    \label{tab:sta-sea_pile_v2}
\end{table}

\begin{table}
    \centering
    \begin{tabular}{lr}
    \hline
    \textbf{Data source}    & \textbf{\# tokens} \\
    \hline
    Internet (common crawl) & 677.6B \\
    Books                   & 56.8B\\
    Academic papers         & 33.6B\\
    Others (QA, etc.)       & 29.6B\\
    Encyclopaedia           & 2.4B\\
    \hline
    Total                   & 800.0B \\
    \hline
    \end{tabular}
    \caption{Number of tokens of the MAP-CC dataset, measured using the original tokenizer from the paper.}
    \label{tab:map-cc}
\end{table}

\begin{table}
    \centering
    \begin{tabular}{lr}
    \hline
    \textbf{Corpus name}    & \textbf{\# tokens} \\
    \hline
    Filtered DCLM           & 752.0B \\
    Decontaminated FLAN     & 17.0B \\
    StackExchange Q\&A      & 1.3B \\
    peS2o                   & 58.6B \\
    Wikipedia/Wikibooks     & 3.7B \\
    Dolmino Math            & 10.7B \\
    \hline
    Total                   & 843.3B \\
    \hline
    \end{tabular}
    \caption{Number of tokens of replay data sampled from OLMo 2's second-stage pretraining, measured using the OLMo 2 tokenizer.}
    \label{tab:olmo2-replay}
\end{table}

\section{Results}
\label{sec:appendix_results}

\begin{table*}
\centering
\setlength{\tabcolsep}{4pt}
\begin{tabular}{ll|cccccccccc|c}
\hline
\textbf{Model} & \textbf{Size} & \textbf{id} & \textbf{km} & \textbf{lo} & \textbf{ms} & \textbf{my} & \textbf{ta} & \textbf{th} & \textbf{tl} & \textbf{vi} & \textbf{zh} & \textbf{Avg} \\
\hline
Mixed & 1B & 42.81 & 21.14 & 23.87 & 42.33 & 4.52 & 12.98 & 24.01 & 32.50 & 29.13 & 23.03 & 25.63 \\
Parallel First & 1B & 35.31 & 13.60 & 18.02 & 35.26 & 3.52 & 11.36	& 20.50 & 32.22 & 23.01 & 15.71 & 20.85 \\
Parallel Last & 1B & 41.00 & 21.46 & 23.76 & 42.02 & 7.29 & 17.71 & 24.35 & 38.31 & 27.99 & 21.59 & 26.55 \\
Multilingual & 1B & 29.61 & 3.07 & 1.17 & 30.10 & 0.30 & 4.07 & 12.98 & 30.53 & 18.24 & 14.96 & 14.50 \\
Parallel Only & 1B & 42.98 & 23.10 & 6.13 & 42.61 & 5.85 & 11.88 & 27.35 & 43.86 & 34.71 & 23.97 & 26.24 \\
\hline
Multilingual & 7B & 41.56 &	26.70 & 30.67 & 41.75 & 10.86 & 21.83 & 27.84 & 44.07 & 34.42 & 21.76 & 30.15\\
Parallel Only & 7B & 49.48 & 32.92 & 34.94 & 49.06 & 22.44 & 33.45 & 35.07 & 51.93 & 41.00 & 30.06 & 38.04 \\
\hline
SeaLLM v3 & 7B & 41.53 & 16.31 & 8.84 & 40.68 & 7.59 & 12.87 & 28.97 & 36.47 & 34.99 & 28.80 & 25.71  \\
Sailor2 & 8B & 46.26 & 32.70 & 37.22 & 46.10 & 24.70 & 28.57 & 32.84 & 48.16 & 39.38 & 27.73 & 36.37 \\
SEA-LION v3.5 & 8B & 42.78 & 26.77 & 26.70 & 41.72 & 20.91 & 26.54 & 30.13 & 42.98 & 35.48 & 26.32 & 32.03\\
\hline
\end{tabular}
\caption{\label{tab:bleu_to_english} BLEU scores for translation into English.}
\end{table*}

\begin{table*}
\centering
\setlength{\tabcolsep}{4pt}
\begin{tabular}{ll|cccccccccc|c}
\hline
\textbf{Model} & \textbf{Size} & \textbf{id} & \textbf{km} & \textbf{lo} & \textbf{ms} & \textbf{my} & \textbf{ta} & \textbf{th} & \textbf{tl} & \textbf{vi} & \textbf{zh} & \textbf{Avg} \\
\hline
Mixed & 1B & 44.99 & 8.61 & 7.75 & 39.83 & 2.18 & 6.21 & 19.65 & 33.45 & 37.93 & 36.32 & 23.69  \\
Parallel First & 1B & 37.44 & 4.58 & 7.83 & 32.97 & 2.26 & 2.51 & 14.60 & 29.54 & 30.04 & 28.19 & 19.00 \\
Parallel Last & 1B & 45.73 & 2.07 & 3.58 & 40.57 & 1.78 & 6.63 & 20.05 & 33.71 & 38.69 & 36.77 & 22.96  \\
Multilingual & 1B & 28.13 & 0.19 & 0.65 & 25.92 & 0.61 & 0.30 & 6.11 & 23.17 & 19.89 & 20.10 & 12.51  \\
Parallel Only & 1B & 45.69 & 3.19 & 6.65 & 40.99 & 2.33 & 8.81 & 21.22 & 34.31 & 39.50 & 38.51 & 24.12 \\
\hline
Multilingual & 7B & 41.40 & 13.87 & 14.61 & 37.66 & 5.28 & 7.73 & 20.62 & 34.30 & 35.30 & 30.90 & 24.17\\
Parallel Only & 7B & 49.50 & 17.99 & 16.47 & 44.50 & 9.25 & 17.58 & 28.66 & 37.92 & 43.44 & 45.86 & 31.12\\
\hline
SeaLLM v3 & 7B & 35.27 & 3.16 & 0.91 & 25.81 & 1.96 & 1.22 & 15.92 & 15.73 & 35.22 & 43.27 & 17.85 \\
Sailor2 & 8B & 47.82 & 22.01 & 22.82 & 41.36 & 15.77 & 11.67 & 27.31 & 37.15 & 43.35 & 41.68 & 31.09 \\
SEA-LION v3.5 & 8B & 42.03 & 12.51 & 12.02 & 36.50 & 8.09 & 12.32 & 21.01 & 31.90 & 39.09 & 38.18 & 25.37 \\
\hline
\end{tabular}
\caption{\label{tab:bleu_from_english} BLEU scores for translation from English.}
\end{table*}

\begin{table*}
\centering
\begin{tabular}{ll|cccc|c}
\hline
\textbf{Model} & \textbf{Size} & \textbf{en} & \textbf{th} & \textbf{vi} & \textbf{zh} & \textbf{Avg} \\
\hline
Mixed & 1B & 51.54 & 41.72 & 44.37 & 35.47 & 43.28 \\
Parallel First & 1B & 52.79 & 43.31 & 44.15 & 39.00 & 44.81 \\
Parallel Last & 1B & 51.00 & 42.91 & 44.15 & 33.29 & 42.84 \\
Multilingual & 1B & 52.77 & 39.42 & 41.86 & 39.08 & 43.28 \\
Parallel Only & 1B & 52.97 & 44.49 & 44.23 & 34.67 & 44.09 \\
\hline
Multilingual & 7B & 52.61 & 45.45 & 47.76 & 37.11 & 45.73\\
Parallel Only & 7B & 52.10 & 47.50 & 47.33 & 34.65 & 45.40\\
\hline
SeaLLM v3 & 7B & 53.77 & 40.32 & 42.87 & 36.17 & 43.28\\
Sailor2 & 8B & 54.65 & 44.45 & 45.55 & 33.69 & 44.59 \\
SEA-LION v3.5 & 8B & 52.79 & 47.05 & 46.23 & 38.36 & 46.11\\
\hline
\end{tabular}
\caption{\label{tab:xnli} XNLI scores for each language.}
\end{table*}

\begin{table*}
\centering
\begin{tabular}{ll|cccccc|c}
\hline
\textbf{Model} & \textbf{Size} & \textbf{en} & \textbf{id} & \textbf{ta} & \textbf{th} & \textbf{vi} & \textbf{zh} & \textbf{Avg} \\
\hline
Mixed & 1B & 77.80 & 63.80 & 53.00 & 56.60 & 63.60 & 59.20 & 62.33   \\
Parallel First & 1B & 79.20 & 64.20 & 54.80 & 54.20 & 63.00 & 58.60 &	 62.33  \\
Parallel Last & 1B & 77.40 & 61.40 & 54.80 & 56.00 & 65.00 & 59.60 & 62.37   \\
Multilingual & 1B & 76.60 & 63.40 & 55.40 & 57.20 & 65.40 & 58.40 & 62.73   \\
Parallel Only & 1B & 77.40 & 61.20 & 54.60 & 55.80 & 61.60 & 59.40 & 61.67  \\
\hline
Multilingual & 7B & 86.00 & 73.60 & 61.60 & 61.60 & 72.80 & 66.20 & 70.30  \\
Parallel Only & 7B & 85.00 & 74.00 & 59.60 & 60.00 & 73.80 & 68.80 & 70.20 \\
\hline
SeaLLM v3 & 7B & 87.80 & 71.00 & 53.80 & 61.20 & 72.40 & 77.00 & 70.53 \\
Sailor2 & 8B & 87.40 & 79.40 & 62.60 & 65.00 & 78.20 & 74.40 & 74.50 \\
SEA-LION v3.5 & 8B & 87.80 & 72.80 & 59.60 & 59.80 & 73.00 & 69.20 & 70.37 \\
\hline
\end{tabular}
\caption{\label{tab:xcopa} XCOPA scores for each language.}
\end{table*}

\begin{table*}
\centering
\begin{tabular}{ll|cc|c}
\hline
\textbf{Model} & \textbf{Size} & \textbf{en} & \textbf{zh} & \textbf{Avg} \\
\hline
Mixed & 1B & 58.80 & 55.80 & 57.30 \\
Parallel First & 1B & 56.90 & 52.85 & 54.88 \\
Parallel Last & 1B & 59.35 & 52.00 & 55.68 \\
Multilingual & 1B & 58.50 & 47.00 & 52.75  \\
Parallel Only & 1B & 61.15 & 53.30 & 57.23\\
\hline
Multilingual & 7B & 66.45 & 53.70 & 60.08\\
Parallel Only & 7B & 70.10 & 59.30 & 64.70\\
\hline
SeaLLM v3 & 7B & 66.85 & 53.50 & 60.18\\
Sailor2 & 8B & 70.90 & 58.15 & 64.53\\
SEA-LION v3.5 & 8B & 66.45 & 60.70 & 63.58\\
\hline
\end{tabular}
\caption{\label{tab:pawsx} PAWS-X scores for each language.}
\end{table*}

In this section, we report the detailed experimental results for each language. We report the models' performance on translating texts from Southeast Asian languages into English (Table~\ref{tab:bleu_to_english}) and English into Southeast Asian languages (Table~\ref{tab:bleu_from_english}). We report the detailed performance on commonsense reasoning tasks, which include XNLI (Table~\ref{tab:xnli}), XCOPA (Table~\ref{tab:xcopa}), and PAWS-X (Table~\ref{tab:pawsx}).

\section{Training Details}
\label{sec:training_details}
We report the architectural configurations of the 1B and 7B models in Table \ref{tab:1B_7B_architecture}, and summarize the corresponding training hyperparameters in Table \ref{tab:training-hparams-1B-7B}. All experiments were conducted using 8× NVIDIA H200 GPUs. Training the 1B model on 10B tokens required approximately 11 hours, while training the 7B model on 34.7B tokens took approximately 180 hours.

\begin{table*}
\centering
\begin{tabular}{lcc}
\hline
\textbf{Hyperparameter} & \textbf{1B} & \textbf{7B} \\
\hline
Number of layers        & 16   & 32   \\
Embedding dimension     & 2048 & 4096 \\
Intermediate dimension  & 8192 & 22016 \\
Attention heads         & 16   & 32   \\
Context window          & 4096 & 4096 \\
Vocabulary size         & 100352 & 100352 \\
\hline
\end{tabular}
\caption{Architecture of the OLMo 2 1B and 7B models.}
\label{tab:1B_7B_architecture}
\end{table*}

\begin{table*}
\centering
\begin{tabular}{lcc}
\hline
\textbf{Hyperparameter} & \textbf{1B} & \textbf{7B} \\
\hline
Global batch size              & 512  & 1024 \\
Micro batch size (per device)  & 4    & 2    \\
Learning rate                  & $2.0\times10^{-4}$ & $2.0\times10^{-4}$ \\
Weight decay                   & 0.1  & 0.1  \\
Optimizer                      & AdamW & AdamW \\
$(\beta_1, \beta_2)$            & (0.9, 0.95) & (0.9, 0.95) \\
Gradient clip                  & 1.0  & 1.0  \\
Max sequence length             & 4096 & 4096 \\
\hline
\end{tabular}
\caption{Training hyperparameters for the 1B and 7B experiments.}
\label{tab:training-hparams-1B-7B}
\end{table*}

\end{document}